\documentclass{article}

\usepackage{microtype}
\usepackage{graphicx}
\usepackage{subcaption}
\usepackage{booktabs} 

\usepackage{hyperref}

\usepackage[accepted]{icml2026}

\usepackage{amsmath}
\usepackage{amssymb}
\usepackage{mathtools}
\usepackage{amsthm}

\usepackage[capitalize,noabbrev]{cleveref}

\theoremstyle{plain}
\newtheorem{theorem}{Theorem}[section]
\newtheorem{proposition}[theorem]{Proposition}
\newtheorem*{proposition*}{Proposition}

\theoremstyle{definition}

\usepackage{xcolor}

\usepackage{xcolor}
\definecolor{indianred}{RGB}{205,92,92}
\definecolor{steelblue}{RGB}{70,130,180}
\definecolor{darkgreen}{RGB}{34,139,34}

\usepackage{newfloat}
\usepackage{listings}
\DeclareCaptionStyle{ruled}{labelfont=normalfont,labelsep=colon,strut=off} 
\floatstyle{ruled}
\newfloat{listing}{tb}{lst}{}
\floatname{listing}{Listing}
\usepackage{booktabs}
\usepackage{multirow}
\usepackage{diagbox}

\usepackage[textsize=tiny]{todonotes}

\icmltitlerunning{Learning Decentralized LLM Collaboration with Multi-Agent Actor Critic}

\begin{document}

\twocolumn[
  \icmltitle{Learning Decentralized LLM Collaboration with Multi-Agent Actor Critic}



  \icmlsetsymbol{equal}{*}

  \begin{icmlauthorlist}
    \icmlauthor{Shuo Liu}{sch}
    \icmlauthor{Tianle Chen}{sch}
    \icmlauthor{Ryan Amiri}{sch}
    \icmlauthor{Christopher Amato}{sch}
  \end{icmlauthorlist}
  \icmlaffiliation{sch}{Northeastern University, Boston, MA}

  \icmlcorrespondingauthor{Christopher Amato}{c.amato@northeastern.edu}

  \icmlkeywords{Multi-Agent Reinforcement Learning, LLM Collaboration, Actor-Critic}

  \vskip 0.3in
]



\printAffiliationsAndNotice{}  

\begin{abstract}
  Recent work has explored optimizing LLM collaboration through Multi-Agent Reinforcement Learning (MARL). However, most MARL fine-tuning approaches rely on predefined execution protocols, which often require centralized execution. Decentralized LLM collaboration is more appealing in practice, as agents can run inference in parallel with flexible deployments. 
  Also, current approaches use Monte Carlo methods for fine-tuning, which suffer from high variance and thus require more samples to train effectively. Actor-critic methods are prevalent in MARL for dealing with these issues; thus, we developed Multi-Agent Actor-Critic (MAAC) methods to optimize decentralized LLM collaboration.
  In this paper, we analyze when and why these MAAC methods are beneficial. We propose 2 MAAC approaches, \textbf{CoLLM-CC} with a \textbf{C}entralized \textbf{C}ritic and \textbf{CoLLM-DC} with \textbf{D}ecentralized \textbf{C}ritics. Our experiments across writing, coding, and game-playing domains show that Monte Carlo methods and CoLLM-DC can achieve performance comparable to CoLLM-CC in short-horizon and dense-reward settings. However, they both underperform CoLLM-CC on long-horizon or sparse-reward tasks, where Monte Carlo methods require substantially more samples and CoLLM-DC struggles to converge. Our code is available at {\small\url{https://github.com/OpenMLRL/CoMLRL/releases/tag/v1.3.6}}.
\end{abstract}


\section{Introduction}

Advanced LLMs have demonstrated remarkable capabilities in natural language understanding and generation \cite{gpt4, gemini, bai2023qwen, deepseek-r1}. This progress has driven growing efforts to transform them into autonomous agents 
\cite{yao2022react, liu2023agentbench, yang2024swe}.

In this context, it is becoming popular to explore coordinating multiple LLMs to improve performance where agents are specified by roles, e.g., generators, planners, or verifiers \cite{autogen, dudebate, skreta2023verifier, chatdev, marti2025, wang2026marti}. Building on the studies of Multi-Agent Systems (MAS) \cite{weiss1999multiagent, peterstonemas, van2008multi, yoavmas}, recent methods employ Multi-Agent Reinforcement Learning (MARL) to optimize their interactions \cite{liu2025spiral, liu2025selfplay, feng2025heterogeneous, chen2025improving, maft, wu2025collabllm, chen2026beyond, feng2026dr}. However, most existing approaches remain confined to predefined execution paradigms, which limits their applicability to broader settings. Moreover, these agents often rely on extensive inter-agent communication to accomplish tasks, requiring centralized execution, which results in limited scalability as well as potential privacy issues in larger-scale MAS.

Decentralized systems have been studied for decades, where each agent is deployed separately and executes independently based on its own observations \cite{waldo1996note, ghosh2006distributed, decpomdp, amato2024introduction}. 
Leveraging decentralized LLMs to complete tasks is beneficial, as it reduces memory and storage pressure on each node and improves the efficiency \cite{huang2019gpipe, lepikhin2020gshard, douillard2023diloco, wu2025deserve}. However, how to effectively optimize these decentralized LLM agents to collaborate remains an open question.

Although Monte Carlo methods are widely adopted in RL fine-tuning due to their simplicity and efficiency \cite{li2023remax, rloo, deepseek-math, deepseek-r1, hureinforcepp, liu2026gdpo}, extending them to optimize multi-LLM collaboration faces many difficulties \cite{liao2025marft, marti2025, hong2025MGRPO, liu2025llmcollaborationmultiagentreinforcement}. Agents need to wait until the end of an episode to receive return signals with high variance. This leads to poor sample efficiency and limits practicality in long-horizon or episodic tasks \cite{rl}.

In this paper, we develop Multi-Agent Actor-critic (MAAC) methods for optimizing decentralized LLM collaboration. We analyze when and why MAAC methods are beneficial for MARL fine-tuning and introduce 2 approaches, \textbf{CoLLM-CC} that employs a centralized critic to estimate joint history values, and \textbf{CoLLM-DC} that uses decentralized critics to estimate individual history values. Our evaluation across writing, coding, and game-playing domains shows that, in dense-reward and short-horizon writing tasks, Monte Carlo methods and CoLLM-DC achieve performance comparable to CoLLM-CC; while in sparse-reward coding tasks and long-horizon Minecraft tasks, both underperform CoLLM-CC, where Monte Carlo methods require substantially more samples for training, and CoLLM-DC fails to converge.

Our contributions are summarized as follows: (i) We develop MAAC methods to optimize decentralized LLM collaboration and analyze their advantages; (ii) We propose CoLLM-DC and CoLLM-CC as 2 MAAC approaches; and (iii) Our results on collaborative writing, coding, and game-playing tasks demonstrate that CoLLM-CC consistently outperforms Monte Carlo methods and CoLLM-DC, particularly in long-horizon tasks with sparse rewards.

\section{Related Works}

\subsection{LLM Collaboration} 

A lot of works employ LLM agents to collaborate with complementary roles (e.g., planner, retriever, generator, verifier) \cite{autogen, dudebate, skreta2023verifier, chatdev, marti2025, liu2025spiral, liu2025selfplay, wu2025collabllm, chen2025improving, zhao2025sirius, zhao2025stronger, feng2026dr, chen2026beyond}. While such role specialization can improve performance, most frameworks enforce rigid protocols, such as refinement pipelines, hierarchical decompositions, or discussion loops, which are hand-coded, entail heavy communication overhead, and typically require centralized execution to produce a solution. Moreover, it is still unclear whether such protocols are optimal cooperation schemes in general \cite{debatefail, cemri2025multi, zhang2025agent}. An alternative is decentralized LLM collaboration, where agents run inference in parallel under limited or no communication \cite{chen2024scalable, qi2025towards, yang2025agentnet, liu2025llmcollaborationmultiagentreinforcement}. This paradigm improves efficiency and enables more flexible deployment across multiple small LLMs. However, how to effectively optimize such decentralized collaboration remains underexplored.

\subsection{Actor-Critic Methods in Cooperative MARL} 

Actor-Critic (AC) is a widely used policy-gradient architecture in which an actor model selects actions based on its policy, and a critic model estimates values during training \cite{rl, actor-critic}. In the decentralized multi-agent learning setting, each agent acts based on its local observation \cite{decpomdp, marl-book}, and the critic can be instantiated either as an individual critic for independent learning \cite{IPPO, lee2022investigation}, or as a centralized critic that conditions on joint history \cite{lowe2017multi, COMA, lyu2021contrasting, MAPPO, lyu2023centralized}. The centralized-critic methods can also leverage additional information during centralized training to support more accurate and stable gradient estimates, while agents still execute in a decentralized manner (CTDE). 

\section{Background}

\subsection{Decentralized LLM Collaboration}

In decentralized LLM collaboration, multiple LLM agents cooperate to solve a class of tasks. Each task is expressed with natural-language observations and provided to the agents as individual prompts. Agents then produce responses in parallel according to their own policies conditioned on the prompt. The responses of all agents are aggregated to form a solution. All agents share a joint reward based on the aggregated outcome. 

After each turn, users, external tools, and other models validate the proposed solution and provide new requirements, suggestions, or constraints for the next iteration. This feedback, together with the dialog history, is maintained and serves as both a decision-making context and an experience for training. The interaction repeats until the task is solved or a predefined turn limit is reached. The goal of agents is to achieve higher-quality solutions with minimal iterations.

Decentralized LLM collaboration offers great advantages by allowing specialized agents to divide and conquer complex problems and operate in parallel to improve efficiency. Rather than relying on a single gigantic LLM in a centralized system, lightweight agents in a decentralized system can focus on subtasks guided by their own prompts, thereby greatly reducing the overhead of maintaining long contexts and extracting relevant semantics. Deploying smaller models locally is also typically easier, safer, and more scalable, making decentralized LLM collaboration well-suited for applications such as long article generation, software engineering, and embodied coordination \cite{gao2023gradientcoin, chen2024scalable, wu2025deserve, liu2026caml,liu2026cross, liu2025llmcollaborationmultiagentreinforcement, hao2025conav, wang2026conavbench}.

\subsection{LLM Dec-POMDP}

Decentralized LLM collaboration can be seen as a subclass of the Dec-POMDP \cite{decpomdp}, which is denoted as $\langle \mathcal{I}, \mathcal{V}, C, M, \mathcal{S}, \{\mathcal{O}_i\}, \{\mathcal{A}_i\}, R, T, H \rangle$.

In an LLM Dec-POMDP, $\mathcal{I}$ denotes a set of $n$ LLM agents, $\mathcal{V}$ is the token vocabulary in LLM inputs and outputs, $C$ is the input context window size for each agent, and $M$ is the max number of tokens in each LLM output. $\mathcal{S}: \mathcal{S}^\mathrm{sys} \times \mathcal{S}^\mathrm{usr}$ denotes the full global state space. At turn $t$, $s_t \in \mathcal{S}$ consists of accessible parts $s^\mathrm{sys}\in \mathcal{S}^\mathrm{sys}$ from the system and inaccessible $s^\mathrm{usr} \in\mathcal{S}^\mathrm{usr}$ parts from users. In RL fine-tuning via verifiable rewards (RLVR), only the accessible parts are used by the reward model. $\mathcal{O}_i$ is the observation space for agent $i$ with $\mathcal{O}=\times_i \mathcal{O}_i$ the joint observation space, where a local observation $o_{i,t}$ is a natural language prompt providing a partial and noisy view of $s_t$, and $|\mathcal{O}_i|=|\mathcal{V}|^C$. $\mathcal{A}_i$ is the action space for agent $i$ with $\mathcal{A}=\times_i \mathcal{A}_i$ the joint action space, where an individual action $a_{i,t}$ is its response to the given prompt, and $|\mathcal{A}_i|=|\mathcal{V}|^M$. The joint reward function $R: \mathcal{S}^\mathrm{sys} \times \mathcal{A} \rightarrow \mathbb{R}$ is implemented via predefined rules or a pretrained reward model. At turn $t$, the joint rewards $r_t \gets R(s^\mathrm{sys}_t, \mathbf{a}_t)$ are determined by the accessible part of current state $s^\mathrm{sys}_t$ and the agents’ joint action $\mathbf{a}_t=\{a_{1, t}, \cdots, a_{n, t}\}$. $T: \mathcal{S} \times \mathcal{A} \rightarrow \Delta(\mathcal{S})$ is the underlying stochastic state transition function. At turn $t$, the agents’ joint actions $\mathbf{a}_t$ induce a shift to a new state $s_{t+1}\sim T(\cdot | s_t, \mathbf{a}_t)$, which reflects the updates in the user state and the states of external models and systems. $H$ is the episode horizon, the turn limit of a dialog.

Since both the accessible $s^\mathrm{sys}$ and the inaccessible $s^\mathrm{usr}$ parts of the states cannot be directly observed by the agents. Each agent maintains its local observation-action history $h_{i,t}=\{o_{i,0}, a_{i,0}, \cdots, o_{i,t}\}$ to infer information about the state $s_t=\{s^\mathrm{sys}, s^\mathrm{usr}\}$. The history of all agents forms a joint history $\mathbf{h}_t = \{h_{1,t}, \cdots, h_{n,t}\}$, and all agents' policies forms a joint policy $\boldsymbol{\pi} =\{\pi_1, \cdots, \pi_n\}$. A solution to a Dec-POMDP is a joint policy that maximizes the expected cumulative reward over the horizon $H$, $\boldsymbol{\pi}^* =\{\pi^*_1, \cdots, \pi^*_n\}=\arg\max_{\boldsymbol{\pi}} \mathbb{E}_{\boldsymbol{\pi}}\left[\sum_{t=0}^{H-1} r_t\right]$. We describe 3 LLM Dec-POMDP instantiations in Section~\ref{sec:exp}.

\section{Cooperative MARL for LLM Fine-Tuning}

\begin{figure*}[t]
    \centering
        \includegraphics[width=0.88\textwidth]{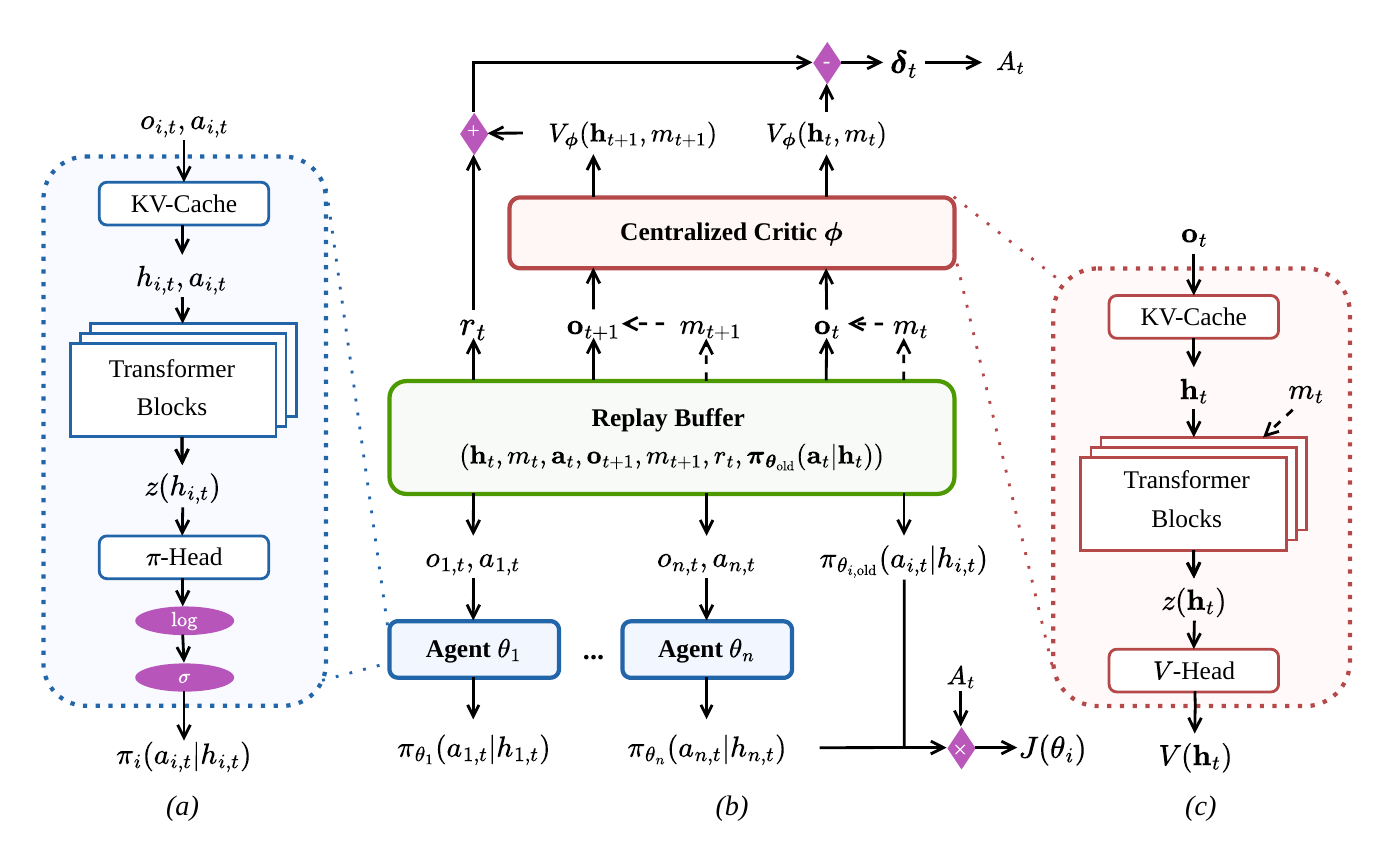}
    \caption{Illustration of CoLLM-CC framework: \textit{(a)} The agent model structure; \textit{(b)} The overall centralized-critic architecture; \textit{(c)} The critic model structure. The corresponding CoLLM-DC framework is shown in Appendix~\ref{app:collm-dc}.}\label{fig:framework}
\end{figure*}

In this section, we introduce cooperative MARL methods and provide a theoretical analysis of their advantages and limitations when applied to LLM fine-tuning.

\subsection{MA-REINFORCE} \label{sec:MA-REINFORCE}

Multi-Agent REINFORCE (MA-REINFORCE) is a class of multi-agent policy gradient methods in which each agent $i$ maintains an individual policy $\pi_{\theta_i}$, and updates it according to Monte-Carlo estimates of joint returns \cite{peshkin2001learning}. Mathematically, 
{\footnotesize
\begin{equation}\label{eq:reinforce}
\begin{gathered}
\nabla_{\theta_i} J(\theta_i)
    = \mathbb{E}_{\boldsymbol{\pi}}\left[
        \sum_{t=0}^{H-1}\rho_{i,t}\,
        \nabla_{\theta_i}\log \pi_{\theta_i}(a_{i,t}\mid h_{i,t})\,
        \big(G(\mathbf{h}_t)-b(\mathbf{h}_t)\big)
    \right],
    \end{gathered}
\end{equation}
}where $G(\mathbf{h}_t)=\mathbb{E}_{\boldsymbol{\pi}}\bigl[\sum_{\tau=t}^{H-1}\gamma^{\tau-t}r_\tau \big| \mathbf{h}_t\bigr]$ is the expected return from $\mathbf{h}_t$, $\rho_{i,t}=\frac{\pi_{\theta_i}(a_{i,t}\mid h_{i,t})}{\pi_{\theta_i, \mathrm{old}}(a_{i,t}\mid h_{i,t})}$ is the importance sampling ratio for correcting the off-policy action sampled, and $b(\mathbf{h}_t)$ denotes an action-independent baseline.
In MA-REINFORCE, agents learn directly from sampled returns without a critic. However, this method can not scale well to long-horizon online learning, because return signals are only available at dialog termination and Monte Carlo estimates suffer from high variance due to accumulated stochasticity. 

MA-REINFORCE is initially designed to take one action at each turn during rollout. Recent MARL fine-tuning methods aim to reduce the variance of gradient estimation by generating $K$ i.i.d. joint actions $\{\mathbf{a}_t^{1}, \cdots, \mathbf{a}_t^{K}\}$ as a group for each history $\mathbf{h}_t$ following policy $\boldsymbol{\pi}(\cdot |\mathbf{h}_t)$ \cite{marti2025, hong2025MGRPO, liu2025llmcollaborationmultiagentreinforcement, feng2026dr}. In the multi-turn environment, each action yields a successor observation $\{\mathbf{o}_{t+1}^{1}, \cdots, \mathbf{o}_{t+1}^{K}\}$, and obtaining lower-variance gradients for these successor histories $\{\mathbf{h}_{t+1}^{1}, \cdots, \mathbf{h}_{t+1}^{K}\}$ also requires $K$ samples. This recursively expands into a $K$-ary rollout tree of depth $H$ \cite{yang2025treerpo, ding2025treegrpo, ji2025tree, li2025treepo}. 

We show that the average gradient estimator by $K$-sampling MA-REINFORCE is unbiased in Proposition~\ref{prop:marflow_unbiased}. 
\begin{proposition}\label{prop:marflow_unbiased}
For each history $\mathbf{h}_t$ at $t$, $t\in[0,H)$, suppose agents sample $K\geq 1$ i.i.d. joint actions $\{\mathbf{a}_t^{1}, \cdots, \mathbf{a}_t^{K}\}$ from $\boldsymbol{\pi}(\cdot| \mathbf{h}_t)$. For each $\mathbf{a}_t^{k}$, an independent $K$-ary rollout tree is produced to estimate the corresponding expected Monte Carlo return $G_t^k$.
For agent $i$, define 
$\bar{g}_{i,t}$ as the averaged gradient estimate over $\{a_{i,t}^{1}, \cdots, a_{i,t}^{K}\}$,
\begin{equation}
\begin{gathered}
    \bar{g}_{i,t}=\frac{1}{K}\sum_{k=1}^K\,\rho_{i,t}^{k}\,\nabla_{\theta_i}\log \pi_{\theta_i}(a^k_{i,t}\mid h_{i,t})\, G_t^k.
\end{gathered}\nonumber
\end{equation}
Then $\bar{g}_{i,t}$ is unbiased for the true gradient $g^*_{i,t}$,
\begin{equation}
\mathbb{E}_{\boldsymbol{\pi}}\!\left[ \bar{g}_{i,t}\mid h_{i,t}\right]
=\mathbb{E}_{\boldsymbol{\pi}}\!\left[g^*_{i,t}\mid h_{i,t}\right]. \nonumber
\end{equation}
\end{proposition}
\begin{proof}[Proof sketch]
The gradient estimates of $K$-sampling MA-REINFORCE equal those of the MA-REINFORCE in \cite{peshkin2001learning}. By Theorem 1 therein, $\bar g_{i,t}$ is unbiased. A full proof is provided in Appendix~\ref{app:proofs}.
\end{proof}

Proposition~\ref{prop:marflow_variance} measures the variance reduction of $ K$-sampling MA-REINFORCE. Under an independent rollout assumption, its variance scales as ${1}/{K^{H-t}}$ in $K$.
\begin{proposition}\label{prop:marflow_variance}
Under the setting of Proposition~\ref{prop:marflow_unbiased}, for any history $h_{i,t}$, we assume the gradient estimates for $\{a_{i,t}^{1}, \cdots, a_{i,t}^{K}\}$ are independent and episodes are not terminated early.
Let $\sigma^2$ be the gradient estimate variance when $K=1$, the variance of $\bar g_{i,t}$ satisfies,
\begin{equation}
\mathrm{Var}_{\boldsymbol{\pi}}(\bar g_{i,t}\mid h_{i,t})=\frac{\sigma^2}{K^{H-t}}.\nonumber
\end{equation}
\end{proposition}
\begin{proof}[Proof sketch]
Without early termination, the $K$-ary rollout tree has $K^{H-t}$ i.i.d. leaf rollouts from $h_{i,t}$. Under independent gradient estimation assumption, all these samples are effectively used in $\bar g_{i,t}$, yielding $\mathrm{Var}(\bar g_{i,t}| h_{i,t})=\sigma^2/K^{H-t}$. We provide the formal justification in Appendix~\ref{app:proofs}.
\end{proof}

However, this variance reduction comes at the cost of a rapidly increasing LLM inference calls to maintain the rollout tree, as shown in Proposition~\ref{prop:marfsampleefficiency}.
\begin{proposition}\label{prop:marfsampleefficiency}
Consider an $H$-horizon episode without early termination $t\in[0,H)$.
Suppose MA-REINFORCE expands a full $K$-ary rollout tree ($K\ge 1$) and, at each history node, draws $K$ i.i.d. joint actions. Each LLM runs one inference at a time to produce a response.
Then the total number of inference calls required for this episode is
\(
N_{\mathrm{call}}(n,K,H)
= \frac{nK(K^H-1)}{K-1}.
\)
\end{proposition}
\begin{proof}[Proof]
Since the rollout forms a $K$-ary tree in each episode, the total number of actions forms a geometric series. Thus, $N_{\mathrm{call}}(n,K,H)= nK\sum_{\ell=0}^{H-1} K^l= \frac{nK(K^H-1)}{K-1}$.
\end{proof}

\subsection{Multi-Agent Actor-Critic} \label{sec:MAAC}

To reduce the variance of gradient estimates and improve sample efficiency, Actor-Critic (AC) methods learn a policy model (actor) $\pi_\theta$ and a value model (critic) $V_\phi$ (or $Q_\psi$). In the multi-agent setting, AC methods are common with Decentralized Critics \cite{COMA, IPPO} or with a Centralized Critic (CC) \cite{lowe2017multi, COMA, MAPPO}.

In DC methods, each agent $i$ maintains a local critic $V_{\phi_i}(h_{i,t})$ and updates its policy $\pi_{\theta_i}(\cdot| h_{i,t})$ via
\begin{equation}
\label{eq:dcpolicy}
\begin{gathered}
\nabla_{\theta_i} J(\theta_i)
=\mathbb{E}_{\boldsymbol{\pi}}\!\left[
\sum_{t=0}^{H-1}\rho_{i,t}\,
\nabla_{\theta_i}\log \pi_{\theta_i}\!\left(a_{i,t}\mid h_{i,t}\right)\,
\delta_{i,t}
\right],
\end{gathered}
\end{equation}
where $\delta_{i,t}
= r_t + \gamma V_{\phi_i}(h_{i,t+1}) - V_{\phi_i}(h_{i,t})$, and each critic $V_{\phi_i}$ is updated by minimizing its TD loss,
\begin{equation}
\label{eq:dcvalue}
    \mathcal{L}(\phi_i)=\mathbb{E}_{\boldsymbol{\pi}}\!\left[\sum_{t=0}^{H-1}\left(r_t+\gamma V_{\phi_i}(h_{i,t+1})-V_{\phi_i}(h_{i,t})\right)^2\right].
\end{equation}

CC learns a shared value function $V_{\boldsymbol{\phi}}(\mathbf{h}_t)$ that conditions on the joint history $\mathbf{h}_t$ (and even other available information during training) to update each agent's policy $\pi_{\theta_i}(\cdot | h_{i,t})$,
\begin{equation}
\label{eq:ccpolicy}
\begin{gathered}
\nabla_{\theta_i} J(\theta_i)
=\mathbb{E}_{\boldsymbol{\pi}}\!\left[
\sum_{t=0}^{H-1}\rho_{i,t}\,
\nabla_{\theta_i}\log \pi_{\theta_i}\!\left(a_{i,t}\mid h_{i,t}\right)\,
\boldsymbol{\delta}_t
\right],
\end{gathered}
\end{equation}
where $\boldsymbol{\delta}_{t}
= r_t + \gamma V_{\boldsymbol{\phi}}(\mathbf{h}_{t+1}) - V_{\boldsymbol{\phi}}(\mathbf{h}_{t})$, and
\begin{equation}
\label{eq:ccvalue}
    \mathcal{L}(\boldsymbol{\phi})=\mathbb{E}_{\boldsymbol{\pi}}\!\left[\sum_{t=0}^{H-1}\left(r_t+\gamma V_{\boldsymbol{\phi}}(\mathbf{h}_{t+1})-V_{\boldsymbol{\phi}}(\mathbf{h}_{t})\right)^2\right].
\end{equation}
Since the critic is just a training construct, it can be removed during execution. As a result, in both DC and CC, agents can still execute in a decentralized manner. 

Assuming convergence of critics, both the gradient estimates of DC in Equation~\ref {eq:dcpolicy} and CC in Equation~\ref{eq:ccpolicy} are unbiased \cite{lyu2021contrasting, lyu2023centralized}. Though this assumption can be harder to satisfy for DC methods, because each individual critic $V_{\phi_i}$ conditions only on its local history $h_{i,t}$, while the policies of the other agents, $\boldsymbol{\pi}_{\boldsymbol{\theta}_{-i}}(\cdot |\mathbf{h}_{-i,t})$, change during training and thus induce non-stationarity.

\section{CoLLM-CC} 

We introduce CoLLM-CC as a representative MAAC approach and discuss key design features, with its DC counterpart, CoLLM-DC, presented in Appendix~\ref{app:collm-dc}.

\subsection{History Representation}

In cooperative MARL, each agent $i$ typically uses a recurrent neural network (RNN) to encode its history $h_{i,t}$ by taking $o_{i,t}$ as input, and its hidden state $z_{i,t}$ serving as a compact but lossy representation \cite{sunehag2017value, COMA, QMIX, QPLEX, baisero2025fixing}. However, this paradigm cannot effectively scale to LLMs that take long prompt sequences as input, and language representations are often high-dimensional. 

Transformers can capture long-range dependencies via attention mechanisms \cite{vaswani2017attention}. This makes them well-suited for modeling dialog and interaction histories and serving as the backbone of modern LLMs. In CoLLM-CC, the dialog history is maintained in a key-value (KV) cache. At each turn, for each agent, we concatenate the KV pairs from previous turns with the new prompt from the environment (\textit{line~5} of Alg.~\ref{alg:collm-cc}), and then maintain only the most recent $C$ pairs in the cache. The agents' KV-caches are independent and private to maintain decentralized inference.
In addition to $\mathbf{h}_t$, the global information $m_t$ can also be incorporated into value estimation to facilitate CC learning a richer or more accurate semantic representation during training. Such global information includes model specifications, task completion progress, or external models or tools existing in the system, which can either be appended as prompts or concatenated with $z(\mathbf{h}_t)$ in a numerical representation. 

\subsection{Sequences as Actions}

Since the human language is highly structured, rewards in many tasks are naturally defined at the level of complete responses. Considering each token an action may lead to uninformative credit assignment. Sequences can be used as macro-actions in RL fine-tuning. However, unlike traditional MARL settings with a small action space and action probabilities are readily available, the response-level action space is combinatorially large $|\mathcal{V}|^M$, making the probability of an entire sequence non-trivial to obtain.

CoLLM-CC employs Teacher-Forced (TF) forward passes to obtain the probability of a response based on the current policy. At dialog turn $t$, for LLM agent $i$, the probability of an response $a_{i,t}$ given $h_{i,t}$ under policy $\pi_{\theta_i}$ factorizes autoregressively as $\pi_{\theta_i}(a_{i,t}| h_{i,t})=\prod_{\mu=1}^M \pi_{\theta_i}(a_{i,t_\mu} | h_{i,t}, a_{i,t_{<\mu}})$,
where $a_{i,t_\mu}$ denotes the $\mu^\mathrm{th}$ token in $a_{i,t}$, and $a_{i,t_{<\mu}}$ is the token prefix up to $\mu$. TF computes the conditional distributions for all target tokens in parallel under a causal mask, and thus the log-probability of the entire sequence can be efficiently obtained in one pass by summing the log-probabilities of each token. TF passes are applied twice in Alg.~\ref{alg:collm-cc}, when agents generate responses to roll out (\textit{line~8}), and when they update their policies (\textit{line~20}).

\begin{algorithm}[t]
\caption{CoLLM-CC}
\begin{algorithmic}[1]
\STATE {\bfseries Input:} Taskset $\mathcal{D}$, LLM agents $\{\pi_{\theta_i}\}_{i\in\mathcal{I}}$, centralized LLM critic $V_{{\boldsymbol{\phi}}}$, learning rates $\alpha_\pi, \alpha_V$, discount $\gamma$, horizon $H$, replay buffer $\mathcal{B}$, training epochs $E$
\FOR{each episode}
    \STATE Sample a task from $\mathcal{D}$
    \STATE Initialize prompts $o_{i,0}$, and $\mathbf{o}_{o}\gets \{o_{i,0}\}_{i=1}^n$
    \STATE Initialize the dialog history $h_{i,0} \gets \{o_{i,0}\}$, $\mathbf{h}_0 \gets \{h_{i,0}\}_{i=1}^n$, and obtain global information $m_0$
    \STATE Initialize replay buffer $\mathcal{B}\leftarrow \emptyset$
    \FOR{turn $t = 0$ to $H-1$}
        \STATE Generate response $a_{i,t} \sim \pi_{\theta_i}(\cdot | h_{i,t})$, and calculate its policy $\pi_{\theta_{i}}(a_{i,t}|h_{i,t})$ with TF, $\forall i \in \mathcal{I}$ 
        \STATE $\mathbf{a}_t\gets\{a_{i,t}\}_{i=1}^n$,  $\boldsymbol{\pi}_{\boldsymbol{\theta}}\gets\{\pi_{\theta_{i}}\}_{i=1}^n$
        \STATE Record behavior policy $\boldsymbol{\pi}_{\boldsymbol{\theta}_\mathrm{old}}\gets\boldsymbol{\pi}_{\boldsymbol{\theta}}$ 
        \STATE Obtain joint reward $r_t$, next turn prompts $o_{i,t+1}$, $\forall i \in \mathcal{I}$, and global information $m_{t+1}$
        \STATE Update $h_{i,t+1} \gets \{h_{i,t}, a_{i,t}, o_{i,t+1}\}$, $\forall i \in \mathcal{I}$
        \STATE $\mathbf{o}_{t+1} \gets \{o_{i, t+1}\}_{i=1}^n$, $\mathbf{h}_{t+1} \gets \{h_{i, t+1}\}_{i=1}^n$
        \STATE Store $\left(\mathbf{h}_t,m_t,\mathbf{a}_t,r_t,\mathbf{o}_{t+1}, m_{t+1},\boldsymbol{\pi}_{\boldsymbol{\theta}_\mathrm{old}}\right)$ into $\mathcal{B}$
    \ENDFOR
    \FOR{training epoch $e=1,\cdots, E$}
        \STATE Sample a minibatch $\beta$ from $\mathcal{B}$
        \FOR{each $\left(\mathbf{h}^b_t,m^b_t,\mathbf{a}^b_t,r^b_t,\mathbf{o}^b_{t+1},m^b_{t+1},\boldsymbol{\pi}^b_{\boldsymbol{\theta}_\mathrm{old}}\right)\in\beta$}
            \STATE Calculate $\mathcal{L}^b(\boldsymbol{\phi},m^b)$ (Eq.~\ref{eq:ccvalue})
            \STATE Calculate $\pi_{\theta_i}(a^b_{i,t}|h^b_{i,t})$ with TF, $\forall i \in \mathcal{I}$
            \STATE Calculate $\nabla_{\theta_i} J^b(\theta_i)$ (Eq.~\ref{eq:ccpolicy}), $\forall i \in \mathcal{I}$
        \ENDFOR
        \STATE Update critic ${\boldsymbol{\phi}} \gets {\boldsymbol{\phi}} - \alpha_V \frac{1}{|\beta|}\sum_{b}\nabla_{\boldsymbol{\phi}} \mathcal{L}^b(\boldsymbol{\phi},m^b)$ 
            \STATE Update actor $\theta_i \gets \theta_i + \alpha_\pi \frac{1}{|\beta|}\sum_{b}\nabla_{\theta_i} J^b(\theta_i)$ 
    \ENDFOR
\ENDFOR
\STATE {\bfseries Output:} $\{\pi_{\theta_i}\}_{i\in\mathcal{I}}$
\end{algorithmic} \label{alg:collm-cc}
\end{algorithm}

\begin{figure*}[t]
    \centering
    \hspace{5mm}
    \begin{subfigure}[b]{0.28\textwidth}
        \centering
        \includegraphics[width=\textwidth]{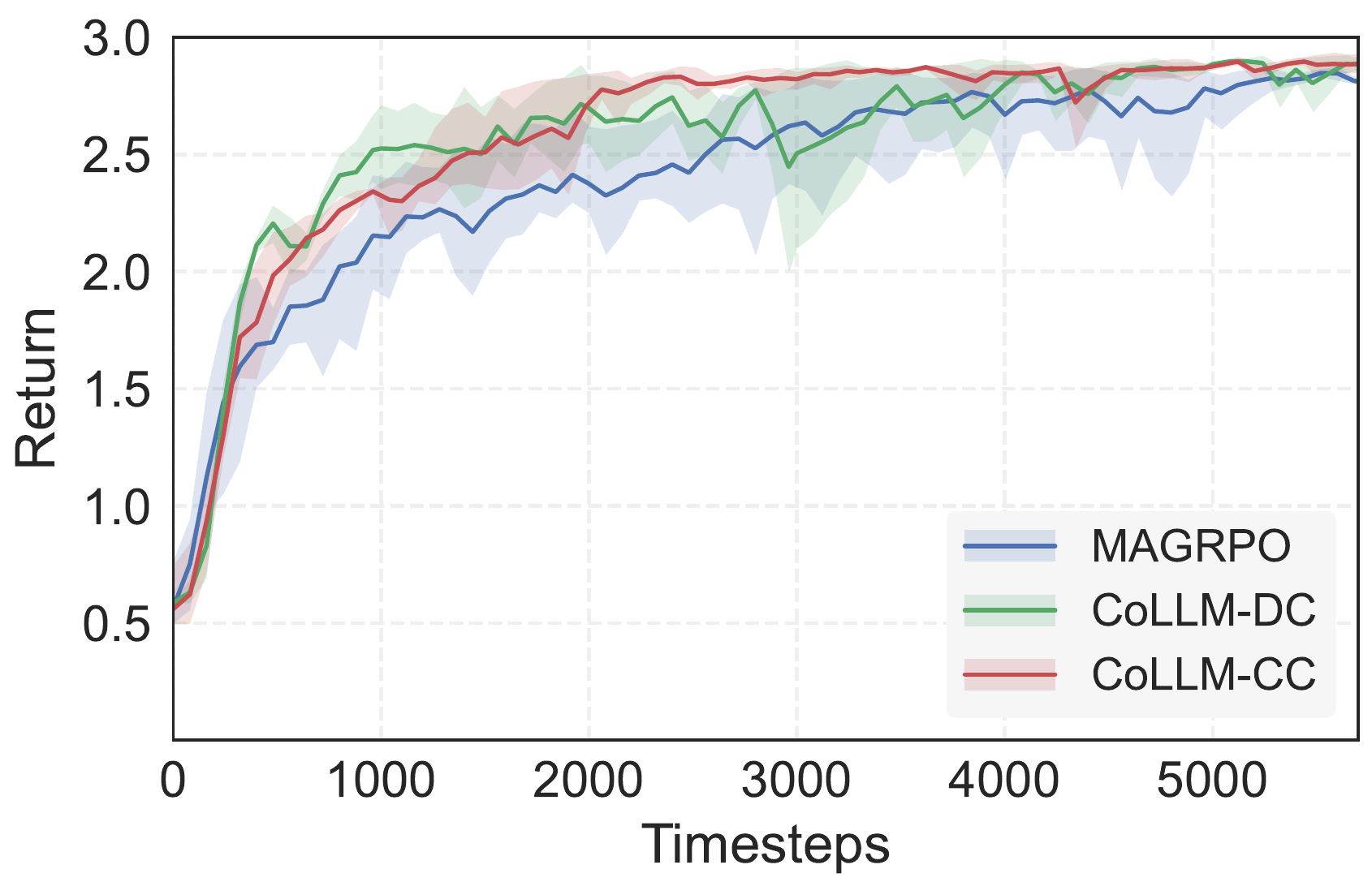}
        \caption{Article Summarization $\mid$ \textit{TLDR}}
        \label{subfig:tldr}
    \end{subfigure}
    \hfill
    \begin{subfigure}[b]{0.28\textwidth}
        \centering
        \includegraphics[width=\textwidth]{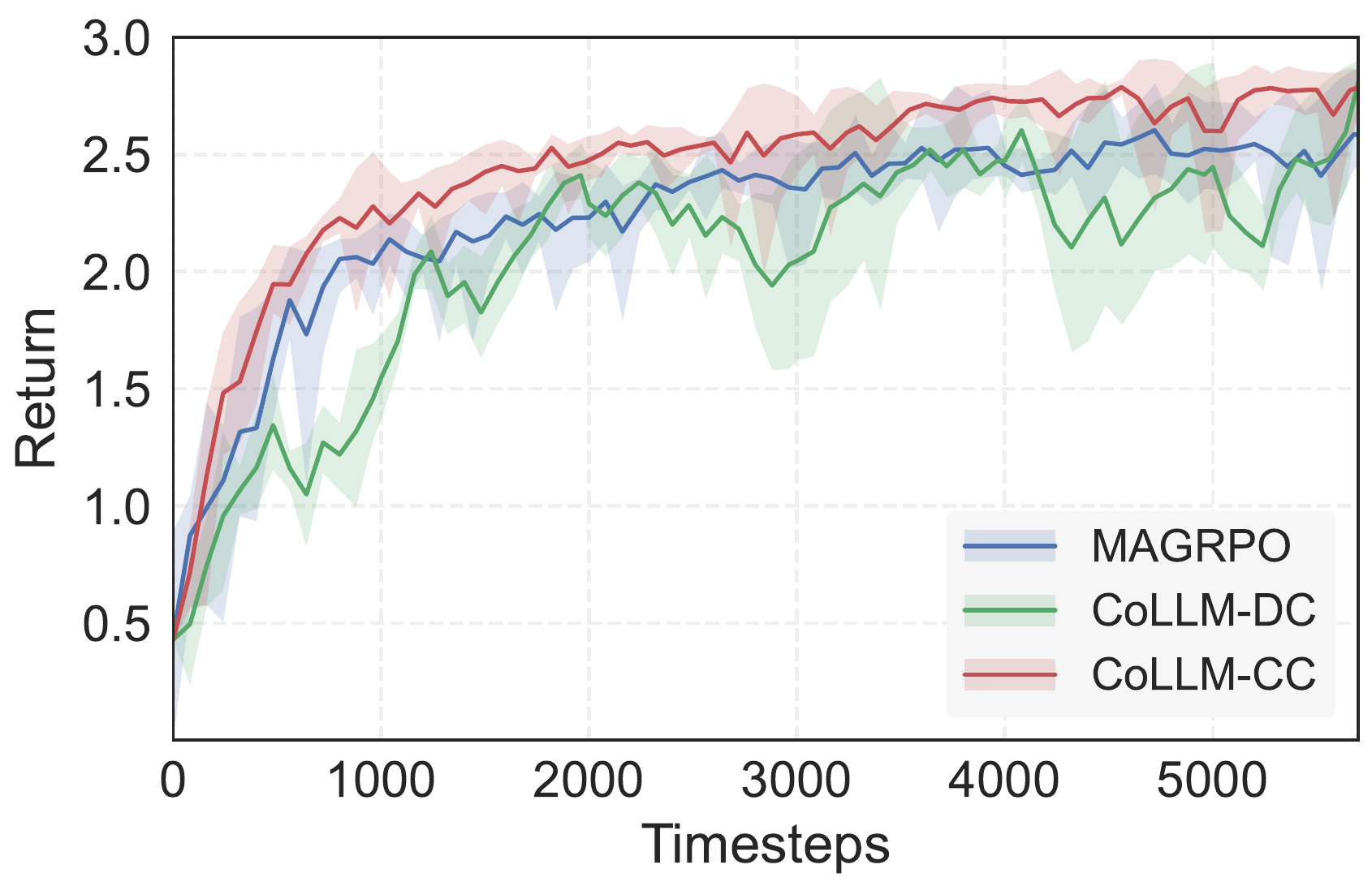}
        \caption{Article Expansion $\mid$ \textit{ArXiv}}
        \label{subfig:arxiv}
    \end{subfigure}
    \hfill
    \begin{subfigure}[b]{0.28\textwidth}
        \centering
        \includegraphics[width=\textwidth]{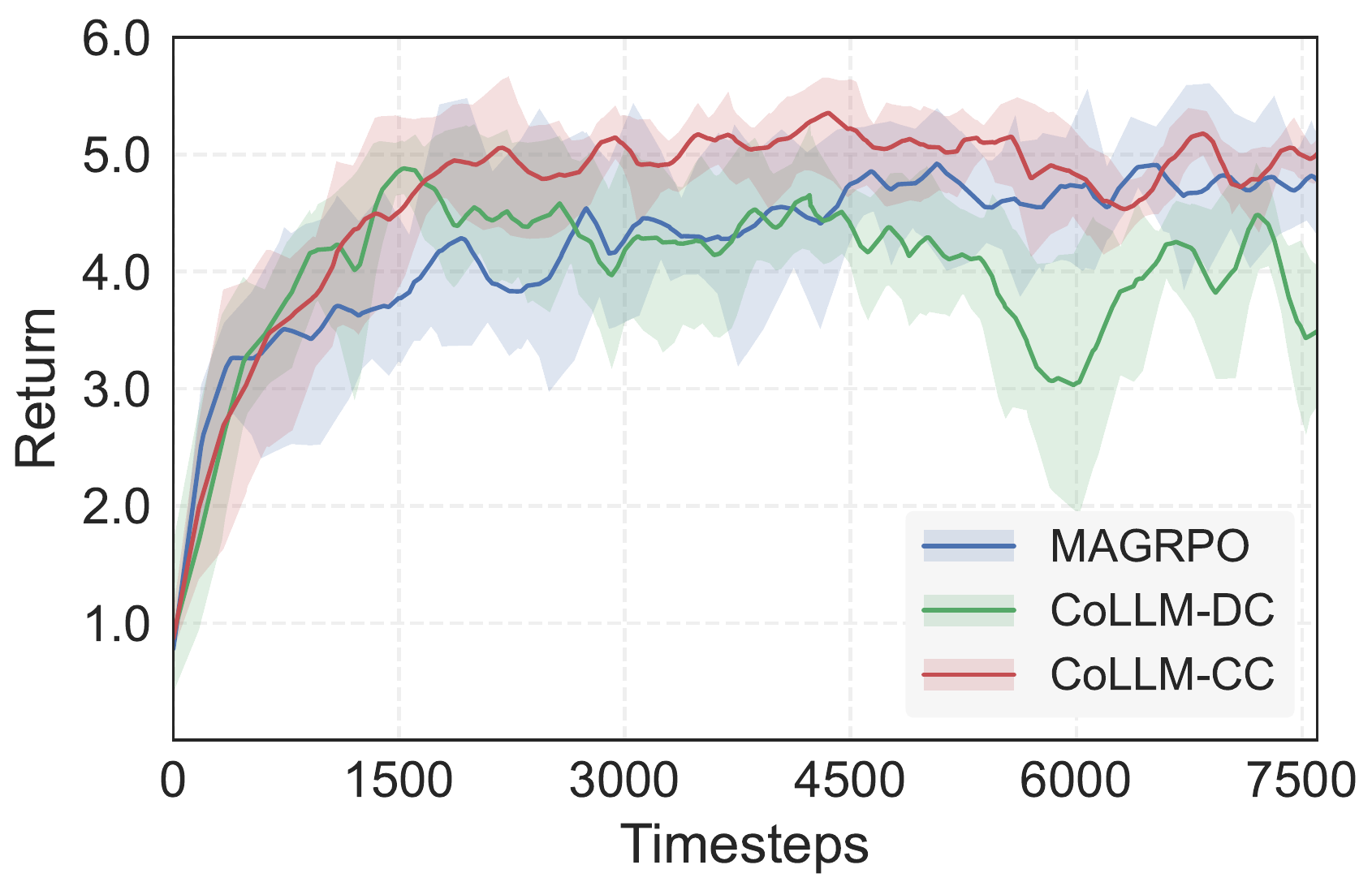}
        \caption{Code Generation $\mid$ \textit{CoopHE}}
        \label{subfig:coophe}
    \end{subfigure}
    \hspace{5mm}
    \vspace{3mm}

    \hspace{85pt}
    \begin{subfigure}[b]{0.28\textwidth}
        \centering
        \includegraphics[width=\textwidth]{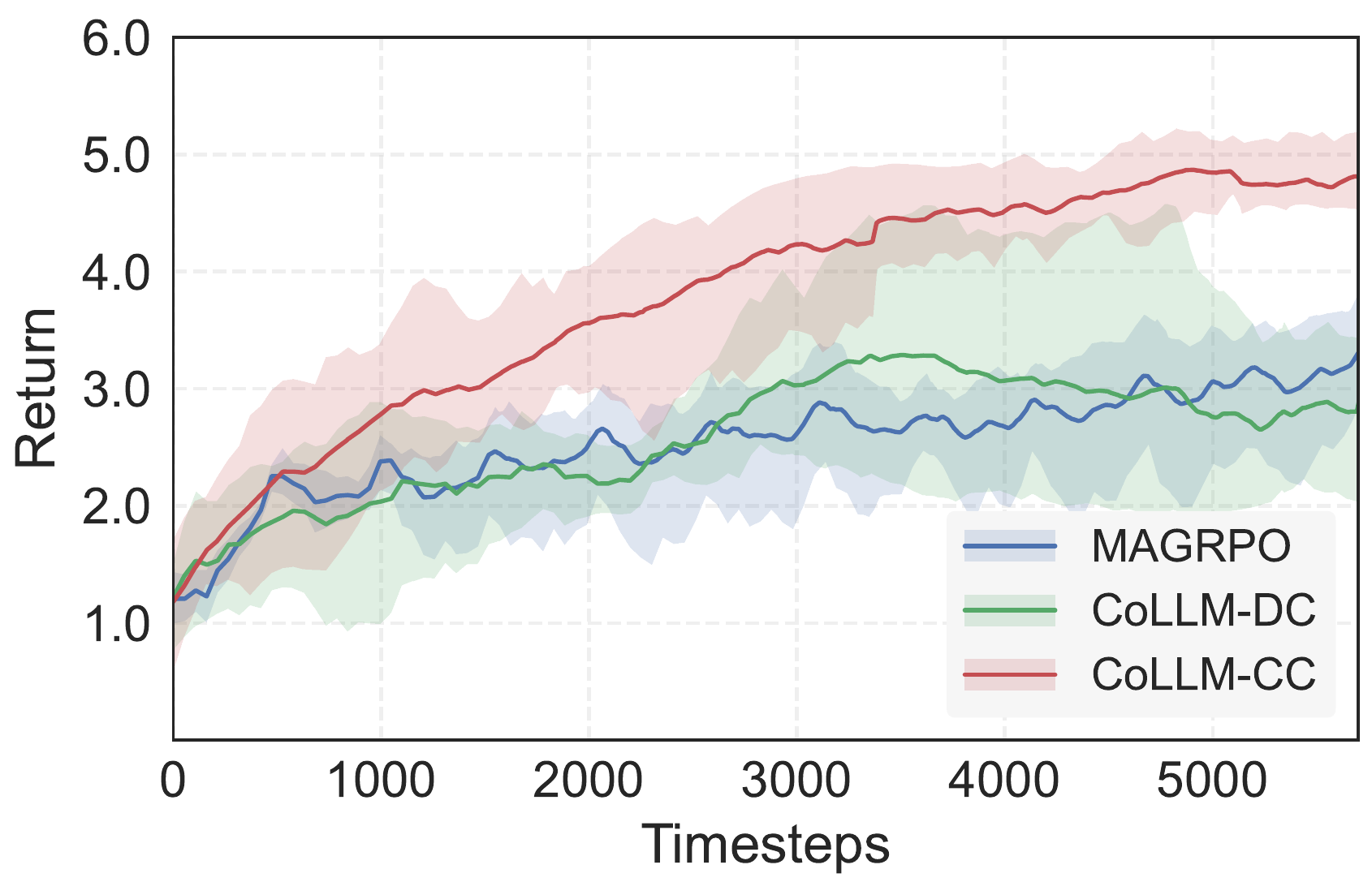}
        \caption{Minecraft Building $\mid$ \textit{StrBuild}}
        \label{subfig:strbuild}
    \end{subfigure}
    \hfill
    \begin{subfigure}[b]{0.28\textwidth}
        \centering
        \includegraphics[width=\textwidth]{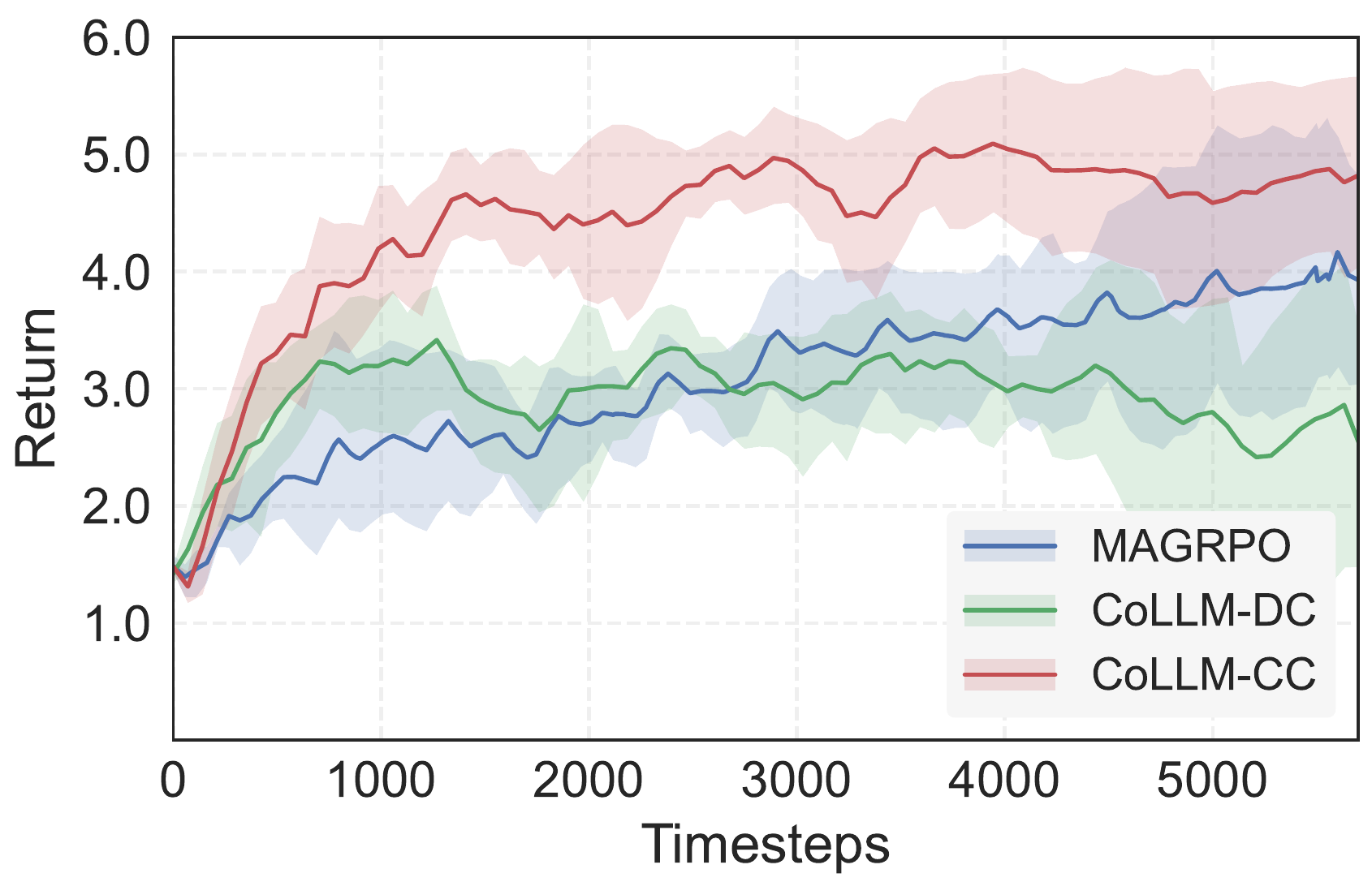}
        \caption{Minecraft Building $\mid$ \textit{HouseBuild}}
        \label{subfig:housebuild}
    \end{subfigure}
    \hspace{85pt}
    \caption{Evaluation results of MAGRPO, CoLLM-DC, and CoLLM-CC across article writing, code generation, and game-playing tasks over 5 runs. The y-axis shows expected return, with limits (min/max) indicating the return scale for each task. Curves are smoothed using a time-weighted exponential moving average. Shaded regions denote 95\% bootstrapped confidence intervals.}
    \label{fig:training}
\end{figure*}

\subsection{Algorithm}

Fig.~\ref{fig:framework}b and Alg.~\ref{alg:collm-cc} illustrate the training procedure of CoLLM-CC.
For each rollout, we first sample a task from $\mathcal{D}$, initialize the agents' prompts as $\mathbf{o}_0\gets \{o_{1, 0}, \cdots, o_{n,0}\}$, and obtain the initial global information $m_0$ from the environment. At each turn $t\in[0,H)$, each agent generates a response $a_{i,t}\sim \pi_{\theta_i}(\cdot | h_{i,t})$ conditioned on its local history $h_{i,t}$. A TF pass is applied right after for each agent to compute the probability $\pi_{\theta_i}(a_{i,t}|h_{i,t})$ of $a_{i,t}$ under $\pi_{\theta_i}(\cdot|h_{i,t})$. All agents' responses form a solution to the task $\mathbf{a}_t\gets \{a_{1,t},\dots,a_{n,t}\}$, and the joint policy can be constructed by $\boldsymbol{\pi}_{\boldsymbol{\theta}}\gets\{\pi_{\theta_i},\cdots,\pi_{\theta_i}\}$. This joint policy is recorded as a behavior policy $\boldsymbol{\pi}_{\boldsymbol{\theta}_\mathrm{old}}(\mathbf{a}_t|\mathbf{h}_t)\gets\boldsymbol{\pi}_{\boldsymbol{\theta}}(\mathbf{a}_t|\mathbf{h}_t)$.

Executing $\mathbf{a}_t$ produces a joint reward $r_t$ from the reward model and triggers the evolution of the environment. The users and systems provide new prompts $o_{i,t+1}$, and the global information is updated as $m_{t+1}$. Each agent updates its history as $h_{i,t+1}\gets \{h_{i,t},a_{i,t},o_{i,t+1}\}$. The resulting transition $(\mathbf{h}_t,m_t,\mathbf{a}_t,r_t,\mathbf{o}_{t+1},\boldsymbol{\pi}_{\boldsymbol{\theta}_\mathrm{old}}(\mathbf{a}_t|\mathbf{h}_t))$, including prompts, global information, responses, reward, new prompts, new global information, and the behavior policy,
is stored into the replay buffer $\mathcal{B}$ for subsequent training. 

At each training epoch, a minibatch $\beta$ of joint transitions is drawn from the replay buffer $\mathcal{B}$. For each sample indexed by $b$, $(\mathbf{h}^b_t,m^b_t,\mathbf{a}^b_t,r^b_t,\mathbf{o}^b_{t+1},m^b_{t+1},\boldsymbol{\pi}^b_{\boldsymbol{\theta}_\mathrm{old}}(\mathbf{a}^b_t|\mathbf{h}^b_t))\in\beta$, the temporal difference $\mathcal{L}^b(\boldsymbol{\phi},m^b_t)$ is calculated according to Equation \ref{eq:ccvalue} by the structure shown in Fig~\ref{fig:framework}c. The probability of outputting $a^b_{i,t}$ by each agent under its current policy $\pi_{\theta_i}(a^b_{i,t}|h^b_{i,t})$ is calculated via a TF pass, which is then used to compute policy gradient $\nabla_{\theta_i} J^b(\theta_i)$ according to Equation~\ref{eq:ccpolicy} in Figure~\ref{fig:framework}a. The gradient of all samples in $\beta$ is averaged to update the centralized critic $\boldsymbol{\phi}$ and all actors $\theta_i$, $\forall i \in \mathcal{I}$, respectively.

\section{Experiments} \label{sec:exp}

We evaluate MAAC on 3 domains: document processing, programming, and language-based games. Experiment settings, additional results, the prompt and reward design, resources used, our code and dataset are provided in Appendix~\ref{app:hyper}, \ref{app:additional-results}, \ref{app:prompt-design}, \ref{app:reward-design}, \ref{app:compute}, \ref{app:codedata}, respectively.

\begin{table*}[t]
\centering
\footnotesize
\caption{Response time (seconds), cost (tokens/agent/turn), and performance (\%) of MAAC and other baselines (single-model, multi-agent test-time interaction, and MA-REINFORCE) on article writing, code generation, and language-based games. Adj denotes the same-texture adjacency rate, and HP denotes the average remaining health points. Response time is measured
on a \textit{NVIDIA GeForce RTX 5090}. \textbf{Bolds} and \underline{underlines} indicate the best performance across all baselines and MARL baselines on each dataset. Results are averaged over 5 runs.}
\setlength{\tabcolsep}{3.5pt}
\begin{tabular}{lccccccccccccccccc}
\toprule
\multirow{2.5}{*}{\textbf{Method}}
& \multicolumn{3}{c}{\textbf{TLDR}}
& \multicolumn{3}{c}{\textbf{arXiv}}
& \multicolumn{3}{c}{\textbf{CoopHE}}
& \multicolumn{4}{c}{\textbf{StrBuild}}
& \multicolumn{4}{c}{\textbf{HouseBuild}} \\
\cmidrule(lr){2-4}
\cmidrule(lr){5-7}
\cmidrule(lr){8-10}
\cmidrule(lr){11-14}
\cmidrule(lr){15-18}
& Time & Cost & Score
& Time & Cost & Score
& Time & Cost & Pass
& Time & Cost & Adj & IoU
& Time & Cost & HP & IoU \\
\midrule
Raw Model            & 5.0 & 465 & 30.3 & 5.1 & 472 & 44.6 & 2.5 & 90  & 56.3 & 10.6 & 427 & 0.9 & 36.6 & 22.6 & 1016 & 99.6 & 43.2 \\
GRPO                 & 4.1 & 387 & 91.7 & 4.2 & 398 & 91.0 & 2.5 & \textbf{88}  & 61.8 & 10.3   & 411  & \textbf{0.4} &  46.1  & 22.0  & 890  & \textbf{100.0} & 54.6   \\
AC                  & 4.0 & 374 & 94.5 & 4.3 & 392 & \textbf{95.3} & 2.5 & 91  & 62.5 & 10.3   & 413  & \textbf{0.4} & 49.8   & 22.1  & 904  & \textbf{100.0} & \textbf{55.9}  \\
\midrule
Parallel             & 2.3 & 244 & 22.9 & 2.3 & 246 & 49.0 & \textbf{\underline{2.3}} & 138 & 50.0 & 9.4  & 232 & 15.7 & 5.9  & 19.2 & 502 & 21.8 & 46.1 \\
Pipeline             & 4.3 & 238 & 21.7 & 3.9 & 203 & 57.8 & 2.6 & 177 & 62.5 & 9.8 & 246 & 12.9 & 18.7 & 20.3 & 488 & 30.6 & 41.3 \\
Discussion           & 4.6 & 234 & 22.3 & 4.8 & 251 & 54.3 & 2.9 & 191 & 25.0 & 10.3 & 236 & 16.2 & 6.5  & 21.0 & 510 & 27.6 & 38.1 \\
\midrule
MAGRPO               & \textbf{\underline{1.8}} & \textbf{\underline{178}} & 93.5 & 2.0 & 201 & 93.1 & \textbf{\underline{2.3}} & 132 & 74.3 & 9.4   & 226  & 13.3 & 50.6 & 19.2   & 446  & 80.2 & 50.9 \\
CoLLM-DC             & 1.9 & 194 & \textbf{\underline{95.4}} & 2.0 & 196 & 94.1 & 2.5 & 161 & 59.1 & \textbf{\underline{9.3}}   & \textbf{\underline{182}}  & 7.6 & 44.6 & 19.4   & 470  & 43.8 & 46.8 \\
CoLLM-CC             & \textbf{\underline{1.8}} & 181 & 95.2 & \textbf{\underline{1.9}} & \textbf{\underline{188}} & \underline{95.0} & 2.6 & \underline{166} & \textbf{\underline{75.2}} & 9.5   & 239  & \underline{7.3} & \textbf{\underline{68.5}} & \textbf{\underline{19.0}}  & \textbf{\underline{442}}  & \underline{86.4} & \underline{52.7} \\
\bottomrule
\end{tabular}
\label{tab:multi_domain_methods}
\end{table*}

\subsection{Setup}

\paragraph{Writing Collaboration}
Processing long documents is time-consuming, whereas parallel execution among decentralized agents can substantially improve efficiency. We frame the common collaborative writing into 2 representative tasks. 
In \textit{TLDR} summarization, 2 \textit{Qwen3-1.7B} agents summarize Reddit posts from the \texttt{prompt} field in \textit{TLDR}. One acts as a high-level summarizer producing a concise paragraph, and one serves as a detailed summarizer providing more comprehensive information. 
In \textit{arXiv} expansion, the same agents expand paper abstracts from the \texttt{abstract} field of \textit{arXiv-public-datasets} into full introductions, where one outlines the research background and motivation, and the other presents the proposed methods and experiments. 
The writing quality is evaluated using a weighted sum of 3 metrics with task-specific hyperparameters. The structure assesses the length ratio of generated responses, encouraging agents to contribute fairly to maximize parallel execution. Style consistency is quantified by a Jaccard similarity score to ensure a uniform tone throughout the article. Logical coherence is measured by a logarithmic function of the frequency of transition words, encouraging natural, smooth transitions between paragraphs.

\paragraph{Coding Collaboration} Leveraging multiple LLM agents for collaborative coding has emerged as a promising paradigm \cite{talebirad2023multi, nvidia_llm_agents, chatdev}.  
We use a code generation task to demonstrate this concept.
To test the collaboration between heterogeneous agents, an auxiliary \textit{Qwen2.5-Coder-3B} assists a primary \textit{Qwen3-4B-Instruct} generator in solving basic programming tasks. The auxiliary agent intends to implement lightweight utilities that support the primary agent to produce the core logic. As many tasks in \textit{HumanEval} and \textit{MBPP} \cite{chen2021evaluating, austin2021program} are atomic and cannot meaningfully benefit from cooperation (e.g., \texttt{add(x,y)}), optimizing collaborative behaviors on such instances introduces substantial noise. Therefore, we construct \textit{CoopHumanEval} dataset (\textit{CoopHE}), which focuses on problems that naturally admit cooperative decomposition. In \textit{CoopHE}, the auxiliary function is named \texttt{aux}, and the main function signature is provided in the \texttt{prompt} field of the problem description.
Agents interact in a multi-turn environment, $H=2$. Agents' outputs are concatenated into a joint program. They receive feedback from static analysis (abstract syntax trees) or dynamic execution (sandbox tests), then proceed to the next turn. The episode terminates if the program passes all tests and the main effectively utilizes the auxiliary function. We use the average pass rate as the evaluation metric. 
Pass@k results are shown in Appendix~\ref{app:additional-results}. 

\begin{figure}[t]
    \centering
    \begin{subfigure}[b]{0.23\textwidth}
        \centering
        \includegraphics[height=2.6cm,width=3.8cm]{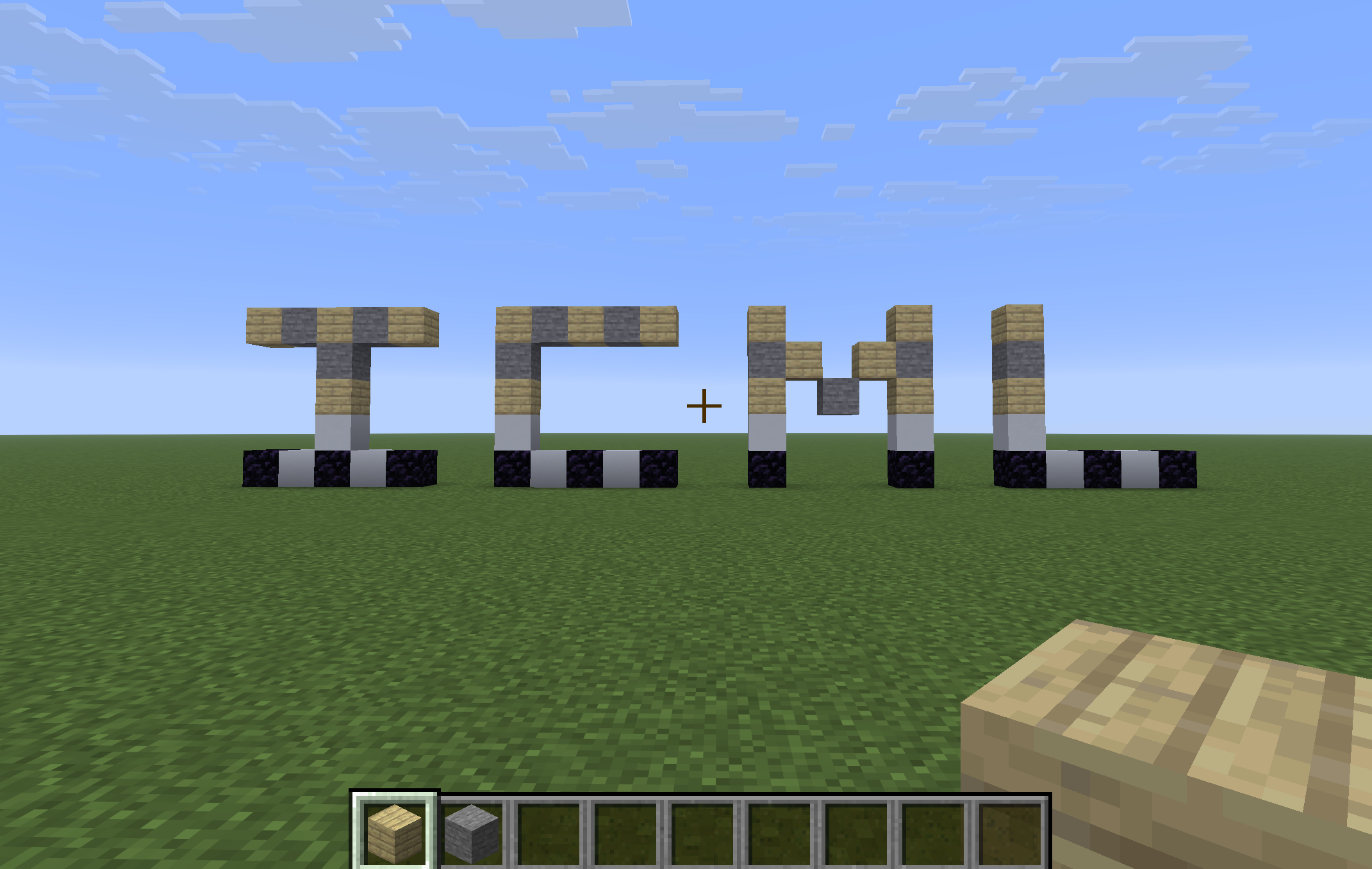}
        \caption{\textit{StrBuild}}
        \label{subfig:MC_2d}
    \end{subfigure}
    \hfill
    \begin{subfigure}[b]{0.23\textwidth}
        \centering
        \includegraphics[height=2.6cm,width=3.8cm]{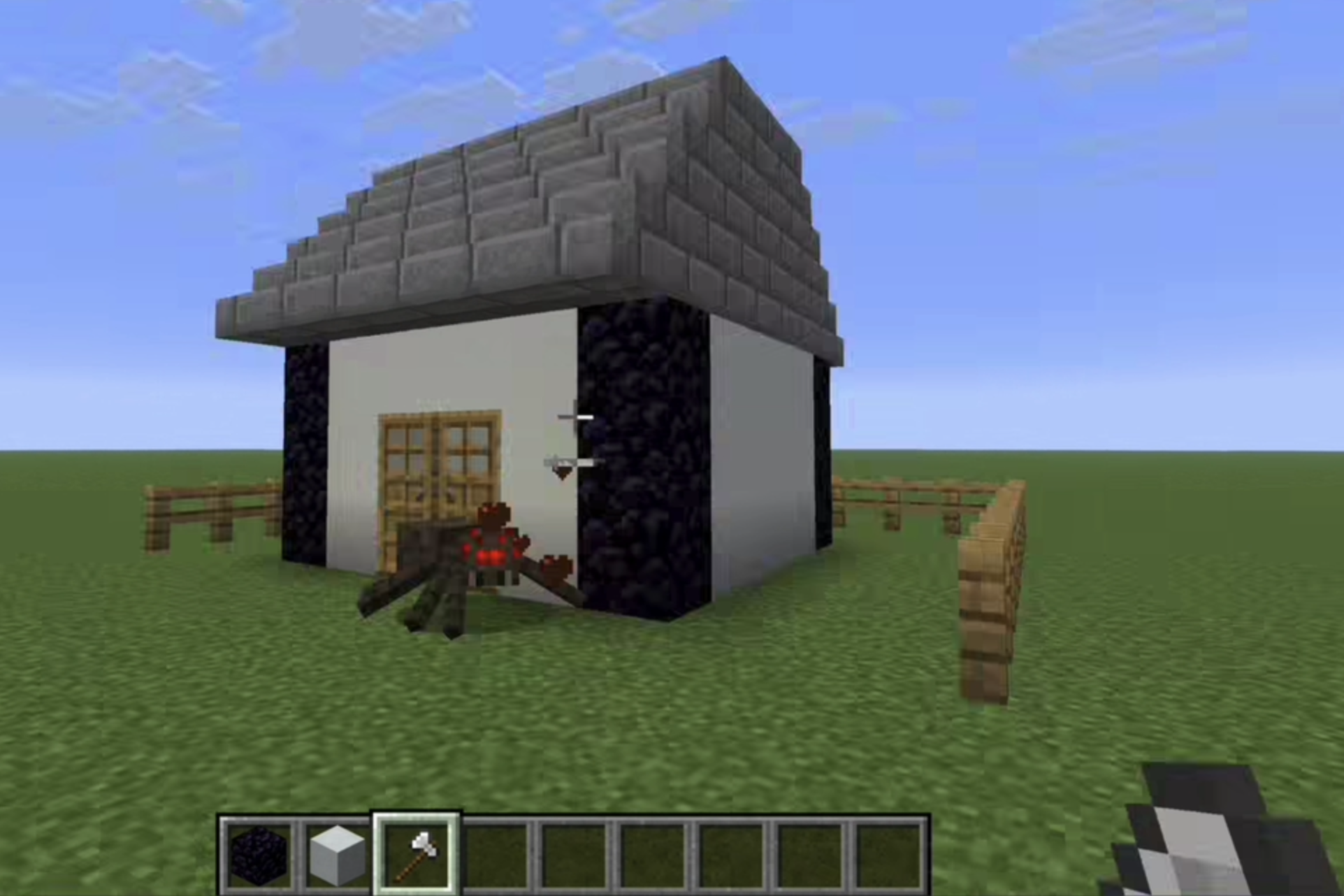}
        \caption{\textit{HouseBuild}}
        \label{subfig:MC_3d}
    \end{subfigure}
    \caption{Screenshots of building tasks in Minecraft. (a) \textit{StrBuild}: The LLM agent with wood outputs a \texttt{/setblock 12 5 5 minecraft:birch\_planks} game instruction to complete the building in ``ICML'' shape. (b) \textit{HouseBuild}: The LLM agent outputs \texttt{/damage @e[type=spider,limit=1] 6 minecraft:player\_attack} to attack a mob, while building a cubic concrete house with a wooden door, 4 obsidian pillars, and a triangular-prism stone roof.\vspace{-2mm}}
    \label{fig:MC_screen}
\end{figure}

\paragraph{Language-based Games}
Language-based games comprise diverse, flexible interactive environments, where agents act and coordinate via language instructions \cite{amongus, multi-agent-bench}.
We consider 2 tasks from \textit{Minecraft} to evaluate LLM collaboration. In \textit{StrBuild}, a \textit{Qwen2.5-3B-Instruct} and a \textit{Qwen3-4B-Instruct} agents collaboratively construct string-like structures (e.g., ``ICML'' in Fig.~\ref{subfig:MC_2d}). Each agent is provided with 2 different block types from wood, stone, concrete, and obsidian, each exhibiting distinct properties, e.g., accessibility and defensive capability. Each agent's resource is limited. The target building is expected to match the exact string while maintaining an even material distribution to enhance structural robustness against potential threats. 
\textit{HouseBuild} involves a higher level of interactivity and stochasticy. The same agents collaboratively construct a house that conforms to predefined architectural specifications (e.g., Fig.~\ref{subfig:MC_3d}) while defending against attacks from hostile mobs (spiders). Since adversaries can actively interfere with agents, effective coordination requires allocating effort between construction and threat mitigation. \textit{StrBuild} and \textit{HouseBuild} unfold in 4 turns, $H=4$. A simulated player gives instructions for incomplete building after each turn. 
For \textit{StrBuild}, we adopt the adjacency rate of same-texture blocks as a task-specific metric, where a lower value corresponds to a more evenly distributed texture across the buildings.
We use the average health point of players (HP) as an indicator of whether the attacks are successfully repelled in \textit{HouseBuild}.
Intersection over Union (IoU) serves as an evaluation metric for both building tasks, with higher values indicating greater compliance with the given specifications. 

\subsection{Baselines}

We evaluate the performance of CoLLM-DC and CoLLM-CC against 3 categories of baselines: single-model methods, multi-agent test-time interaction, and other MARL methods.

\paragraph{Single-Model Methods}
Single-model baselines contain the results from a larger base model or its fine-tuned variants. We select a comparable-size model as the sum of MAS: \textit{Qwen3-4B-Instruct} for writing tasks, \textit{Qwen2.5-7B} \textit{Coder} and \textit{Instruct} models for code generation and Minecraft game-playing, respectively. The reward model is the same as that used in MARL methods, except for metrics specifically designed for cooperation. We set all single-model baselines to be single-turn, as the external feedback for single-model and multi-agent approaches differ significantly.

\paragraph{Multi-Agent Test-Time Interaction Methods}
We compare our method against 3 prompt-based multi-agent interaction baselines, where the same agents operate only through prompts \cite{autogen}. In the parallel baseline, agents act independently without explicit communication. The sequential pipeline baseline represents a one-way communication setting where agents act in turn and can later observe the inputs and outputs of earlier agents. The discussion baseline frames the multi-agent debate \cite{dudebate}, where agents engage in two-way, asynchronous communication. Each agent can access the others' outputs in the next turn and generate its response accordingly. To make a fair comparison, we set the number of message-passing rounds to the same across multi-agent methods and keep the prompt adjustment minimal.

\paragraph{MARL Methods}
We compare CoLLM-DC and CoLLM-CC with a representative MA-REINFORCE approach (Section~\ref{sec:MA-REINFORCE}), Multi-Agent Group Relative Policy Optimization (MAGRPO) \cite{liu2025llmcollaborationmultiagentreinforcement}. In MAGRPO, the group relative baseline $b^\mathrm{grp}(\mathbf{h}_t)=\frac{1}{K}\sum_{k=1}^k G_{t}^k$ is the average return for $K$ i.i.d.\ responses $\{\mathbf{a}^1,\cdots,\mathbf{a}^K\}$ on each joint history $\mathbf{h}_t$. We set $K=4$ for short-horizon writing and coding tasks, and $K=2$ for Minecraft tasks due to the rapid increase in the number of required samples (Proposition~\ref{prop:marfsampleefficiency}). The same learning rate is applied to update each sample across all 3 methods to ensure a fair comparison of sample efficiency (Appendix~\ref{app:additional-results}). We use the task completion progress and the current turn number as $m_t$ in both CC and DC for consistency, though such information is often unavailable to DC in standard MARL. 

\begin{table*}[t]
    \centering
    \caption{
        Performance of MARL algorithms for decentralized LLM collaboration on Minecraft tasks with $n=2$ and $n=3$ agents. The $n=2$ setting, marked with $\dagger$, serves as the control group. For $n=3$, subscripts report the numerical change relative to the corresponding $n=2$ result; arrows indicate whether the metric value increases or decreases, while colors indicate whether the change improves (\textcolor{indianred}{red}) or degrades (\textcolor{steelblue}{blue}) performance. Results are averaged over 5 runs. The best results among the three methods are marked in \textbf{bold}.
    }
    \label{tab:ablation-more-agents}
    \resizebox{0.8\textwidth}{!}{
    \begin{tabular}{lcccccccc}
        \toprule
        \multirow{2}{*}{\textbf{Method}}
        & \multicolumn{2}{c}{\textbf{StrBuild} $(n=2)$}
        & \multicolumn{2}{c}{\textbf{StrBuild} $(n=3)$}
        & \multicolumn{2}{c}{\textbf{HouseBuild} $(n=2)$}
        & \multicolumn{2}{c}{\textbf{HouseBuild} $(n=3)$} \\
        \cmidrule(lr){2-3}
        \cmidrule(lr){4-5}
        \cmidrule(lr){6-7}
        \cmidrule(lr){8-9}
        & Adj & IoU
        & Adj & IoU
        & HP & IoU
        & HP & IoU \\
        \midrule
        MAGRPO
        & $13.3^{\dagger}$ & $50.6^{\dagger}$
        & $8.7_{\textcolor{indianred}{\,\downarrow 4.6}}$  & $38.1_{\textcolor{steelblue}{\,\downarrow 12.5}}$
        & $80.2^{\dagger}$ & $50.9^{\dagger}$
        & $84.0_{\textcolor{indianred}{\,\uparrow 3.8}}$ & $46.3_{\textcolor{steelblue}{\,\downarrow 4.6}}$ \\
        CoLLM-DC
        & $7.6^{\dagger}$  & $44.6^{\dagger}$
        & $9.4_{\textcolor{steelblue}{\,\uparrow 1.8}}$  & $34.5_{\textcolor{steelblue}{\,\downarrow 10.1}}$
        & $43.8^{\dagger}$ & $46.8^{\dagger}$
        & $50.6_{\textcolor{indianred}{\,\uparrow 6.8}}$ & $43.7_{\textcolor{steelblue}{\,\downarrow 3.1}}$ \\
        CoLLM-CC
        & $\mathbf{7.3}^{\dagger}$  & $\mathbf{68.5}^{\dagger}$
        & $\mathbf{6.9}_{\textcolor{indianred}{\,\downarrow 0.4}}$  & $\mathbf{75.4}_{\textcolor{indianred}{\,\uparrow 6.9}}$
        & $\mathbf{86.4}^{\dagger}$ & $\mathbf{52.7}^{\dagger}$
        & $\mathbf{94.2}_{\textcolor{indianred}{\,\uparrow 7.8}}$ & $\mathbf{61.3}_{\textcolor{indianred}{\,\uparrow 8.6}}$ \\
        \bottomrule
    \end{tabular}
    }
\end{table*}

\subsection{Evaluation Results} 

The performance comparison between MARL methods and other baselines is presented in Table~\ref{tab:multi_domain_methods}, where the evaluation curves of MAGRPO (\textcolor{steelblue}{blue}), CoLLM-DC (\textcolor{darkgreen}{green}), and CoLLM-CC (\textcolor{indianred}{red}) are presented in Fig.~\ref{fig:training}.

\paragraph{Writing Collaboration} In writing tasks, single-model methods achieve strong performance but have low inference speed and high cost due to their larger model size. Since the raw models are not specifically optimized for coordination-centric objectives, prompt-based multi-agent methods produce low-quality solutions. Both RL and MARL fine-tuning guide agents to generate concise, high-quality content, leading to shorter response times, lower token usage, and improved performance.
MAGRPO and CoLLM-CC achieve similar performance after convergence, indicating that $K=4$ samples are sufficient for accurate value estimation. MAGRPO has slightly higher variance and lower sample efficiency, as shown by a larger shaded region and slower convergence in Fig.~\ref{subfig:tldr} and \ref{subfig:arxiv}. CoLLM-DC struggles to achieve stable convergence in \textit{arXiv} expansion, as DC based on local information cannot provide stationary signals for gradient estimates.

\paragraph{Coding Collaboration} In coding collaboration, fine-tuning \textit{Qwen2.5-Coder-7B} yields limited improvement ($\leq 6.2\%$), as coding logics vary substantially across training and test problems and no general strategy consistently improves pass rates across diverse tasks. Parallel execution and one-round discussion perform poorly, as agents lack timely information to accurately assess the correctness and functionality of others’ functions. Sequential generation achieves performance comparable to the single-model baseline but incurs substantially higher inference latency because agents must wait for others’ responses before proceeding.
With $K=4$, MAGRPO achieves comparably high pass rates $74.3\%$ as CoLLM-CC $75.2\%$. The main generator provides fallback solutions for potential vulnerabilities in the auxiliary, and it effectively uses external feedback to correct errors. However, MAGRPO requires significantly more samples to converge in Fig.~\ref{subfig:coophe}, reaching stability at approximately $5000$ timesteps, compared to $2000$ timesteps for CoLLM-CC. CoLLM-DC exhibits oscillations in later stages ($4500$ timesteps) and thus cannot perform well, with only $59.1\%$ pass rate. This instability is caused by sparse rewards in code generation, where a single incorrect token can invalidate functionality and induce abrupt reward drops.

\paragraph{Minecraft Games} In Minecraft games, the fine-tuned single model achieves the lowest same-texture adjacency rate $0.4$-$0.9\%$ in \textit{StrBuild} and the full health points in \textit{HouseBuild}. This is because the agents do not need to infer others’ material choices or attack behavior, thus the task is relatively easier. Prompt-based multi-agent methods perform particularly poorly in these domains, likely because the raw models \textit{Qwen2.5-3B-Instruct} and \textit{Qwen3-4B-Instruct} are not specifically optimized with Minecraft building instructions. CoLLM-CC achieves the best IoU $68.5\%$ in \textit{StrBuild} by leveraging external hints and outperforms all methods except the single model fine-tuned by AC in \textit{HouseBuild} with IoU $52.7\%$.
In Fig~\ref{subfig:strbuild} and \ref{subfig:housebuild}, MAGRPO and CoLLM-DC perform substantially worse than CoLLM-CC. Due to the rapid growth of the rollout tree, a smaller sampling $K=2$ is needed, resulting in slower convergence and higher-variance updates, as discussed in Appendix~\ref{app:additional-results}. CoLLM-DC fails to converge on these tasks since non-stationarity accumulates across $4$ turns, leading to more unstable value estimates.

\paragraph{Results in Larger-Scale MAS}
We also evaluate our methods in a larger-scale setting with $n=3$ agents. We keep the reward design unchanged and introduce an additional agent with the same prompt structure. The results are shown in Table~\ref{tab:ablation-more-agents}. Compared with the $n=2$ setting, all methods achieve higher HP on \textit{HouseBuild}, suggesting that additional agents can provide better defense against spiders. On \textit{StrBuild}, MAGRPO achieves lower Adj $\downarrow 4.6\%$, but this comes at the cost of less accurate adherence to the building specification, i.e., lower IoU $\downarrow 12.5\%$. Moreover, MAGRPO incurs substantially longer training time (Appendix~\ref{app:additional-results}), as suggested in Proposition~\ref{prop:marfsampleefficiency}. CoLLM-DC also shows worse performance when scaling to more agents, with $\uparrow1.8\%$ Adj and $\downarrow10.1\%$ on \textit{StrBuild} and $\downarrow3.1\%$ on \textit{HouseBuild}. This is because each decentralized critic observes only a limited portion of the global state, making the non-stationarity issue more severe. In contrast, CoLLM-CC achieves the strongest overall performance when $n=3$, with both lower Adj $\downarrow 0.4\%$, higher HP $\uparrow7.8\%$, and IoU $\uparrow6.9\%$ and $\uparrow8.6\%$ on \textit{StrBuild} and on \textit{HouseBuild}, respectively. These results indicate that CoLLM-CC can effectively support cooperative learning in larger-scale MAS under decentralized execution.

\section{Conclusion}

Decentralized LLM collaboration accelerates inference and enables flexible deployment, making it promising. We developed centralized critic (CoLLM-CC) and decentralized critic (CoLLM-DC) methods to optimize cooperation among decentralized LLMs. Evaluations on writing, coding, and game-playing tasks show that MARL-based methods can achieve equal or better performance than a single larger model.
Among MARL methods, MA-REINFORCE and CoLLM-DC struggle in long-horizon, sparse-reward settings due to low sample efficiency and convergence issues, while CoLLM-CC achieves the best performance with the lowest variance and highest sample efficiency.

\paragraph{Limitations} 
Our work has several limitations. First, although our MAAC methods (CoLLM-DC and CoLLM-CC) can in principle be applied to online learning, we do not extend our experiments to longer-horizon tasks. This is because MAGRPO has low sample efficiency, making fair comparisons prohibitively expensive in such settings. Extending these methods to more complex agentic domains remains an important direction for future work. Second, as discussed in Appendix~\ref{app:collm-dc}, more efficient variants of CoLLM-DC are possible. However, these variants may introduce additional challenges, such as gradient conflicts. It remains worth exploring whether these issues can be mitigated, and whether similar efficiency improvements can be incorporated into centralized-critic methods. Finally, our methods focus on a strictly decentralized setting with concurrent execution and no inter-agent communication. Designing methods under relaxed communication or coordination constraints remains an open question.

\section*{Acknowledgment}

This work was partially funded by NSF grants  
\#2409351 and \#2525087. It used Delta and DeltaAI computing resources
at the National Center for Supercomputing Applications
through allocation CIS251326 from the Advanced Cyberinfrastructure Coordination Ecosystem: Services \& Support program, which is supported by NSF grants
\#2138259, \#2138286, \#2138307, \#2137603, and \#2138296. 

This work was completed in part using the Explorer Cluster, supported by Northeastern University’s Research Computing team. Shuo Liu received computing support from Lambda’s Research Grant Program.

\section*{Impact Statement}

This paper presents work whose goal is to advance the field of machine learning. There are many potential societal consequences of our work, none of which we feel must be specifically highlighted here.

\newpage
\bibliography{ref}
\bibliographystyle{icml2026}

\newpage
\appendix
\twocolumn

\section{Proofs} \label{app:proofs}

Here we provide the full proofs for the propositions in our paper.

\subsection{Proof of Proposition~\ref{prop:marflow_unbiased}}

\begin{proposition*}
For each history $\mathbf{h}_t$ at $t$, $t\in[0,H)$, suppose agents sample $K\geq 1$ i.i.d. joint actions $\{\mathbf{a}_t^{1}, \cdots, \mathbf{a}_t^{K}\}$ from $\boldsymbol{\pi}(\cdot| \mathbf{h}_t)$. For each $\mathbf{a}_t^{k}$, an independent $K$-ary rollout tree is produced to estimate the corresponding expected Monte Carlo return $G_t^k$.
For agent $i$, define 
$\bar{g}_{i,t}$ as the averaged gradient estimate over $\{a_{i,t}^{1}, \cdots, a_{i,t}^{K}\}$,
\begin{equation}
\begin{gathered}
    \bar{g}_{i,t}=\frac{1}{K}\sum_{k=1}^K\,\rho_{i,t}^{k}\,\nabla_{\theta_i}\log \pi_{\theta_i}(a^k_{i,t}\mid h_{i,t})\, G_t^k.
\end{gathered}\nonumber
\end{equation}
Then $\bar{g}_{i,t}$ is unbiased for the true gradient term $g^*_{i,t}$,
\begin{equation}
\mathbb{E}_{\boldsymbol{\pi}}\!\left[ \bar{g}_{i,t}\mid h_{i,t}\right]
=\mathbb{E}_{\boldsymbol{\pi}}\!\left[g^*_{i,t}\mid h_{i,t}\right]. \nonumber
\end{equation}
\end{proposition*}

\begin{proof}
We show that the expectation of gradient estimates $\bar g_{i,t}$ of $K$-sampling MA-REINFORCE ($K \geq 2$) is equal to $g^1_{i,t}$ of MA-REINFORCE with $K=1$.

For agent $i$ at time $t$, define
\[
g_{i,t}^k =
\rho_{i,t}^k
\nabla_{\theta_i} \log \pi_{\theta_i}(a_{i,t}^k \mid h_{i,t})
\, G_t^k.
\]

By linearity of expectation,
{\footnotesize
\begin{equation}
\begin{aligned}
    \mathbb{E}_{\boldsymbol{\pi}}\!\left[ G_t^k \mid h_{i,t}^k \right]
&=
\mathbb{E}_{\boldsymbol{\pi}}\!\left[ \frac{1}{K} \sum_{\ell=1}^K G_{t+1}^\ell \mid h_{i,t}^k \right]\\
&=
\frac{1}{K} \sum_{\ell=1}^K\mathbb{E}_{\boldsymbol{\pi}}\!\left[ G_{t+1}^\ell \mid h_{i,t}^k \right],
\end{aligned}\nonumber
\end{equation}
}
where $\ell$ denotes the sampling index at $t+1$. Since at the terminal step $t=H-1$, $G_{H-1}^k=r_{H-1}^k$, and $g_{i,H-1}^k$ is unbiased. We can derive backward that $g_{i,t}^k$ is unbiased for all $t$ and $k$. Thus, according to Theorem 1 of \cite{peshkin2001learning}, $\bar g_{i,t}=g_{i,t}^1$ is unbiased.
\end{proof}

\subsection{Proof of Proposition~\ref{prop:marflow_variance}}

\begin{proposition*}
Under the setting of Proposition~\ref{prop:marflow_unbiased}, for any history $h_{i,t}$, we assume the gradient estimates for $\{a_{i,t}^{1}, \cdots, a_{i,t}^{K}\}$ are independent and episodes are not terminated early.
Let $\sigma^2$ be the gradient estimate variance when $K=1$, the variance of $\bar g_{i,t}$ satisfies,
\begin{equation}
\mathrm{Var}_{\boldsymbol{\pi}}(\bar g_{i,t}\mid h_{i,t})=\frac{\sigma^2}{K^{H-t}}.\nonumber
\end{equation}

\end{proposition*}

\begin{proof}
We show that, when the episodes are not terminated early and expand into a full $K$-ary tree, and assuming gradient estimates are independent for $\{a_{i,t}^{1}, \cdots, a_{i,t}^{K}\}$, the variance of $g_{i,t}^K$ scales inversely with $K$.

For agent $i$ at time $t$, define
\[
g_{i,t}^k =
\rho_{i,t}^k
\nabla_{\theta_i} \log \pi_{\theta_i}(a_{i,t}^k \mid h_{i,t})
\, G_t^k.
\]

Then, by variance decomposition, we have
{\footnotesize
\[
\begin{aligned}
&\quad\mathrm{Var}\!\left(\bar{g}_{i,t} \mid h_{i,t}\right)\\
&=\mathrm{Var}\!\left(\frac{1}{K}\sum_{k=1}^K g^k_{i,t} \mid h_{i,t}\right)\\
&= \frac{1}{K^2} \sum_{k=1}^{K} \mathrm{Var}\!\left(g^k_{i,t} \mid h_{i,t}\right) + \frac{2}{K^2} \sum_{1 \le k < l \le K}
\mathrm{Cov}\!\left(g^k_{i,t}, g^l_{i,t} \mid h_{i,t}\right)
\end{aligned},
\]
}
where $k$ and $l$ are different i.i.d. samples at $t$. Since $\{a_{i,t}^{1}, \cdots, a_{i,t}^{K}\}$ are i.i.d., we have $\mathrm{Cov}(g^k_{i,t}, g^l_{i,t} | h_{i,t})=0$. Assuming the episode is not terminated early, at $t+1$, we have $\ell$ new rollouts for each $a_{i,t}^{k}$, and so on, we have $\mathrm{leaf}=K^{H-1-t}$ leaf nodes and returns at $t=H-1$.

{\footnotesize
\[
\begin{aligned}
&\quad\mathrm{Var}\!\left(\bar{g}_{i,t} \mid h_{i,t}\right)\\
&= \frac{1}{K} \mathrm{Var}\!\left(g^k_{i,t} \mid h_{i,t}\right)\\
&= \frac{1}{K} \mathrm{Var}\!\left(\rho_{i,t}^k
\nabla_{\theta_i} \log \pi_{\theta_i}(a_{i,t}^k \mid h_{i,t})
\, G_t^k \mid h_{i,t}^k  \right)\\
&=\frac{1}{K} \mathrm{Var}\!\left(\rho_{i,t}^k
\nabla_{\theta_i} \log \pi_{\theta_i}(a_{i,t}^k \mid h_{i,t})
\, \frac{1}{K} \sum_{\ell=1}^{K} G_{t+1}^\ell\mid h_{i,t}^k  \right)\\
&=\frac{1}{K} \mathrm{Var}\!\left(\rho_{i,t}^k
\nabla_{\theta_i} \log \pi_{\theta_i}(a_{i,t}^k \mid h_{i,t})
\, \frac{\sum_{\mathrm{leaf}=1}^{K^{H-1-t}} G_{H-1}^\mathrm{leaf}}{K^{H-1-t}} \mid h_{i,t}^k  \right)\\
&= \frac{1}{K}\frac{1}{(K^{H-1-t})^2} \sum_{\mathrm{leaf}=1}^{K^{H-1-t}} \mathrm{Var}\!\left(\rho_{i,t}^k
\nabla_{\theta_i} \log \pi_{\theta_i}(a_{i,t}^k \mid h_{i,t})
\, G_{t}^\mathrm{leaf} \mid h_{i,t}^k  \right)\\
&\qquad + \frac{2}{K}\frac{2}{(K^{H-1-t})^2} \sum_{1\leq \mathrm{leaf\_1}\leq\mathrm{leaf\_2}\leq K}^{K^{H-1-t}} \,\mathrm{Cov}\!\left(g_{i,t}^\mathrm{leaf\_1}, g_{i,t}^\mathrm{leaf\_2} \mid h_{i,t}^k  \right)
\end{aligned}
\]
}

Assuming the $\{g_{i,t}^{1}, \cdots, g_{i,t}^{K}\}$ and are independent,\footnote{However, this independent gradient assumption is usually hard to satisfy in Dec-POMDP, because dialog histories with the same prefixes usually have positively correlated values.} $\mathrm{Cov}(g_{i,t}^\mathrm{leaf\_1}, g_{i,t}^\mathrm{leaf\_2} | h_{i,t}^k)=0$. 

Thus, we have,
\[
\begin{aligned}
&\quad\mathrm{Var}\!\left(\bar{g}_{i,t} \mid h_{i,t}\right)\\
&= \frac{1}{K^{H-t}} \mathrm{Var}\!\left(\rho_{i,t}^k
\nabla_{\theta_i} \log \pi_{\theta_i}(a_{i,t}^k \mid h_{i,t})
\, G_{t}^\mathrm{leaf} \mid h_{i,t}^k  \right)\\
&= \frac{\sigma^2}{K^{H-t}} 
\end{aligned}
\]

\end{proof}

\newpage
\section{CoLLM-DC} \label{app:collm-dc}

\begin{figure*}[t]
    \centering
        \includegraphics[width=0.95\textwidth]{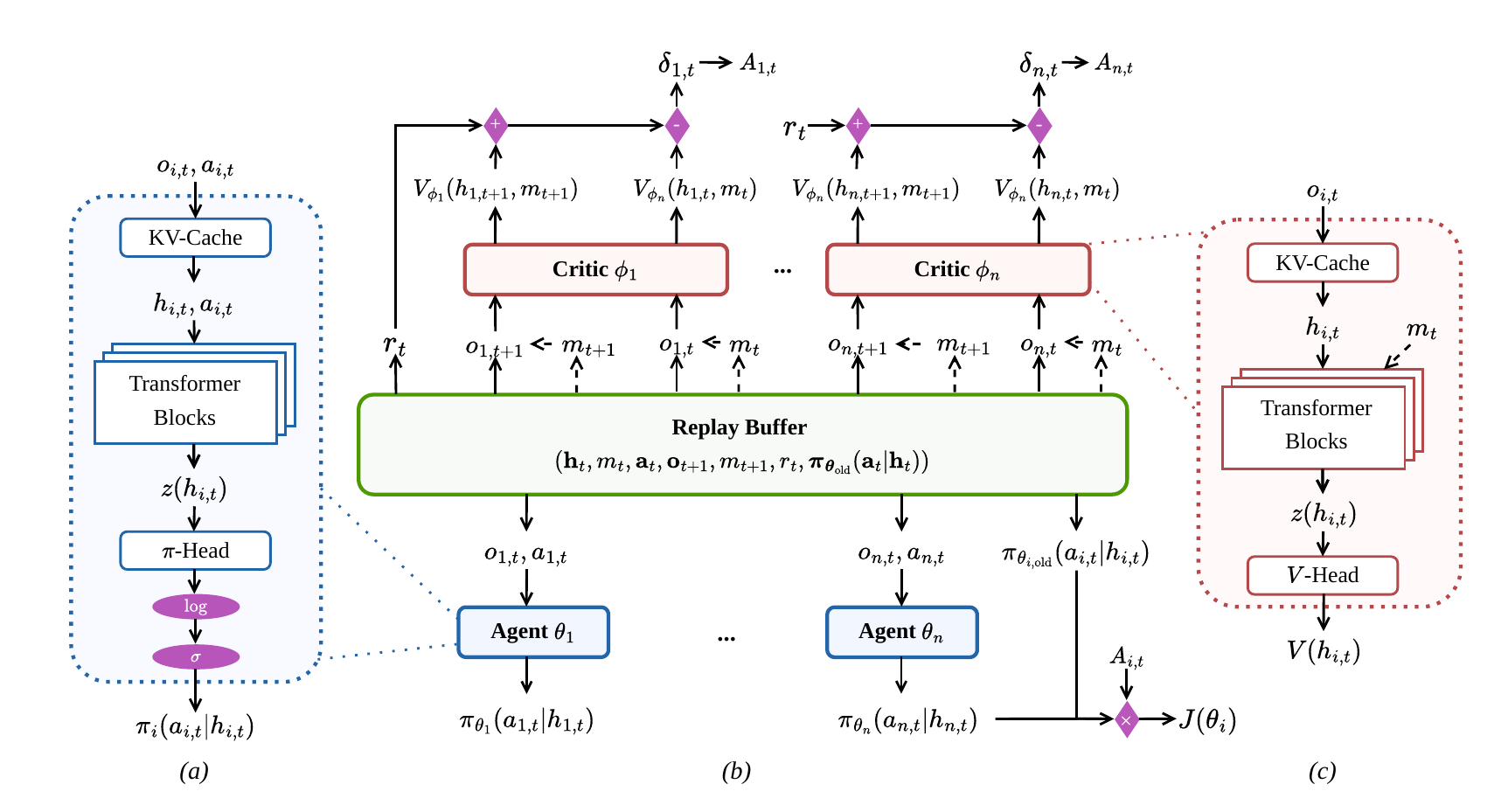}
    \caption{CoLLM-DC framework: \textit{(a)} The agent structure; \textit{(b)} The overall decentralized-critic architecture; \textit{(c)} The critic structure.}\label{fig:framework-dc}
    \vspace{-1mm}
\end{figure*}

\begin{algorithm}[H]
\caption{CoLLM-DC}
\begin{algorithmic}[1]
\STATE {\bfseries Input:} Taskset $\mathcal{D}$, LLM agents $\{\pi_{\theta_i}\}_{i\in\mathcal{I}}$, decentralized LLM critics $\{V_{\phi_i}\}_{i\in\mathcal{I}}$, learning rates $\alpha_\pi, \alpha_V$, discount $\gamma$, horizon $H$, replay buffer $\mathcal{B}$, training epochs $E$
\FOR{each episode}
    \STATE \fbox{\textit{Roll out same as CoLLM-CC in Alg.~\ref{alg:collm-cc} (line 3-15)}}
    \FOR{training epoch $e=1,\cdots, E$}
        \STATE Sample a minibatch of joint transitions $\beta$ from $\mathcal{B}$
        \FOR{each agent $i \in \mathcal{I}$}
            \FOR{each sample agent $i$'s transition $\beta_i$, $(h^b_{i,t},m^b_t, a^b_{i,t},r^b_t,o^b_{i,t+1},m^b_{t+1}\pi^b_{\theta_i,\mathrm{old}}(a^b_{i,t}|h^b_{i,t}))$ $\in\beta_i$}
                \STATE Calculate TD loss $\mathcal{L}^b_i(\phi_i)$ (Eq.~\ref{eq:dcvalue})
                \STATE Calculate $\pi_{\theta_i}(a^b_{i,t}|h^b_{i,t})$ with TF
                \STATE Calculate $\nabla_{\theta_i} J^b(\theta_i)$ (Eq.~\ref{eq:dcpolicy})
            \ENDFOR
            \STATE Update critic $\phi_i \gets \phi_i - \alpha_V \frac{1}{|\beta|}\sum_{b}\nabla_{\phi_i}\mathcal{L}^b_i(\phi_i)$
            \STATE Update actor $\theta_i \gets \theta_i + \alpha_\pi \frac{1}{|\beta|}\sum_{b}\nabla_{\theta_i}J^b(\theta_i)$
        \ENDFOR
    \ENDFOR
\ENDFOR
\STATE {\bfseries Output:} $\{\pi_{\theta_i}\}_{i\in\mathcal{I}}$
\end{algorithmic} \label{alg:collm-dc}
\end{algorithm}

Fig.~\ref{fig:framework-dc} and Alg.~\ref{alg:collm-dc} show the training procedure of CoLLM-DC. The rollout and actor updates are the same as those in CoLLM-CC (Alg.~\ref{alg:collm-cc}). The only difference is in the training phase. In each training epoch, a minibatch $\beta_i$ of joint transitions is drawn from $\mathcal{B}$. For each sample,
$(h^b_{i,t},m^b_t, a^b_{i,t},r^b_t,o^b_{i,t+1},m^b_{t+1}\pi^b_{\theta_i,\mathrm{old}}(a^b_{i,t}|h^b_{i,t}))\in\beta_i$,
each agent $i$ computes its own TD loss $\mathcal{L}^b_i(\phi_i)$ using a decentralized critic conditioned only on its local history, following the structure shown in Fig.~\ref{fig:framework-dc}c. The probability $\pi_{\theta_i}(a^b_{i,t}| h^b_{i,t})$ under the current policy is computed by a TF pass and is used to calculate the policy gradient $\nabla_{\theta_i} J^b(\theta_i)$ according to Eq.\ref{eq:dcpolicy} in Fig.\ref{fig:framework-dc}a. The gradients of all samples in $\beta_i$ are averaged to update each agent’s critic $\phi_i$ and actor $\theta_i$ independently, $\forall i \in \mathcal{I}$.

It is noteworthy that the CoLLM-DC architecture in Fig.~\ref{fig:framework-dc} has an efficient variant based on parameter sharing. Unlike CoLLM-CC, the agents in Fig.~\ref{fig:framework-dc}\textit{a} and Fig.~\ref{fig:framework-dc}\textit{c} learn the same latent representation $z(h_{i,t})$, and thus can potentially share a single model to accelerate learning. However, because the optimization objectives in Eq.~\ref{eq:dcvalue} and \ref{eq:dcpolicy} are different, such parameter sharing may introduce gradient interference during training.

\section{Experimental Settings} \label{app:hyper}

We introduce the experimental settings in Fig.~\ref{fig:training} and Table~\ref{tab:multi_domain_methods}.

\subsection{Datasets}

We list the preprocessed datasets we used for our training.

\begin{itemize}
    \item \textit{\textbf{Writing Collaboration}}
    \begin{itemize}
        \item Training set:
        \begin{itemize}
            \item \textit{TLDR}[0:1000]
            \item \textit{arXiv}[0:1000]
        \end{itemize}
    \item Test set:
        \begin{itemize}
            \item \textit{TLDR}[1000:1100]
            \item \textit{arXiv}[1000:1100]
        \end{itemize}
    \end{itemize}
    
    \item \textit{\textbf{Code Collaboration}}
    \begin{itemize}
    \item Training set: \textit{CoopHE}[0:66]
    \item Test set: \textit{CoopHE}[66:82]
    \end{itemize}

    \item \textit{\textbf{Minecraft Game-Playing}}
    \begin{itemize}
    \item Training set:
        \begin{itemize}
            \item \textit{StrBuild}[0:8]
            \item \textit{HouseBuild}[0:8]
        \end{itemize}
    \item Test set:
        \begin{itemize}
            \item \textit{StrBuild}[8:10]
            \item \textit{HouseBuild}[8:10]
        \end{itemize}
    \end{itemize}
\end{itemize}

\subsection{Architectures}

We list the architectures of the agent and critic models used in our training.

\begin{itemize}
    \item \textit{\textbf{Writing Collaboration}}
    \begin{itemize}
        \item Agents
        \begin{itemize}
            \item \textit{Qwen3-1.7B}
            \item \textit{Qwen3-1.7B}
        \end{itemize}
        \item Critic (if applicable): \textit{Qwen3-1.7B}
        \item Temperature: 0.7
        \item Top-$p$: 0.9
        \item Top-$k$: null
        \item Max output tokens: 256
    \end{itemize}

    \item \textit{\textbf{Code Collaboration}}
    \begin{itemize}
        \item Agents
        \begin{itemize}
            \item \textit{Qwen2.5-Coder-3B}
            \item \textit{Qwen3-4B-Instruct-2507}
        \end{itemize}
        \item Critic (if applicable): \textit{Qwen2.5-Coder-3B}
        \item Temperature: 0.6
        \item Top-$p$: 0.6
        \item Top-$k$: null
        \item Max output tokens: 256
    \end{itemize}

    \item \textit{\textbf{Minecraft Game-Playing}}
    \begin{itemize}
        \item Agents
        \begin{itemize}
            \item \textit{Qwen2.5-3B-Instruct}
            \item \textit{Qwen3-4B-Instruct-2507}
        \end{itemize}
        \item Critic (if applicable): \textit{Qwen3-4B-Instruct-2507}
        \item Temperature: 0.6
        \item Top-$p$: 0.6
        \item Top-$k$: null
        \item Max output tokens
        \begin{itemize}
            \item \textit{StrBuild}: 256
            \item \textit{HouseBuild}: 512
        \end{itemize}
    \end{itemize}
\end{itemize}

\subsection{Hyperparameters}

We show the key hyperparameters used in MAGRPO, CoLLM-DC, and CoLLM-CC.

\begin{itemize}
    \item \textit{\textbf{Writing Collaboration}}
    \begin{itemize}
        \item Number of turns: 1
        \item Number of generations: 4
        \item Rollout buffer size: 4

        \item Number of train epochs
            \begin{itemize}
                \item MAGRPO: 2
                \item CoLLM-DC/CoLLM-CC: 20
            \end{itemize}

        \item Agent learning rate: $5 \times 10^{-6}$
        \item Critic learning rate (if applicable): $3 \times 10^{-6}$
        \item Advantage clip: 0.2
        \item Number of evaluation samples: 4
    \end{itemize}

    \item \textit{\textbf{Code Collaboration}}
    \begin{itemize}
        \item Number of turns: 2
        \item Number of generations: 4
        \item Rollout buffer size
            \begin{itemize}
                \item MAGRPO: 16
                \item CoLLM-DC/CoLLM-CC: 4
            \end{itemize}
        \item Number of train epochs
            \begin{itemize}
                \item MAGRPO: 8
                \item CoLLM-DC/CoLLM-CC: 80
            \end{itemize}

        \item Agent learning rate
        \begin{itemize}
            \item MAGRPO: $2 \times 10^{-5}$
            \item CoLLM-DC/CoLLM-CC: $5 \times 10^{-6}$
        \end{itemize}

        \item Critic learning rate (if applicable): $3 \times 10^{-6}$
        \item Advantage clip: 0.2
        \item Number of evaluation samples: 4
    \end{itemize}

    \item \textit{\textbf{Minecraft Game-Playing}}
    \begin{itemize}
        \item Number of turns: 4
        \item Number of generations: 2
        \item Rollout buffer size
            \begin{itemize}
                \item MAGRPO: 2
                \item CoLLM-DC/CoLLM-CC: 1
            \end{itemize}

        \item Number of train epochs
        \begin{itemize}
            \item MAGRPO: 16
            \item CoLLM-DC/CoLLM-CC: 120
        \end{itemize}

        \item Agent learning rate
        \begin{itemize}
            \item MAGRPO:
            \begin{itemize}
                \item \textit{StrBuild}: $5 \times 10^{-6}$
                \item \textit{HouseBuild}: $1 \times 10^{-5}$
            \end{itemize}
            \item CoLLM-DC/CoLLM-CC:
            \begin{itemize}
                \item \textit{StrBuild}: $2.5 \times 10^{-6}$
                \item \textit{HouseBuild}: $5 \times 10^{-6}$
            \end{itemize}
        \end{itemize}

        \item Critic learning rate (if applicable):
        \begin{itemize}
            \item CoLLM-DC/CoLLM-CC:
            \begin{itemize}
                \item \textit{StrBuild}: $1.5 \times 10^{-6}$
                \item \textit{HouseBuild}: $3 \times 10^{-6}$
            \end{itemize}
        \end{itemize}
        \item Advantage clip: 0.05
        \item Number of evaluation samples: 2
    \end{itemize}
\end{itemize}

\section{Additional Results} \label{app:additional-results}

We provide additional results in our experiments.

\subsection{Pass@k Results on CoopHE}

\begin{table}[H]
\centering
\footnotesize
\setlength{\tabcolsep}{3pt}
\begin{tabular}{lcccc}
\toprule
\textbf{Method} & \textbf{Pass@1} & \textbf{Pass@3} & \textbf{Pass@5} & \textbf{Pass@10} \\
\midrule
Single-Agent & 56.3 & 58.7 & 61.9 & 62.5 \\
GRPO         & 61.8 & 62.0 & 62.2 & 62.8 \\
AC          & 62.5 & 62.6 & 63.0 & 63.3 \\
\midrule
Parallel     & 50.0 & 51.1 & 68.8 & 68.8 \\
Pipeline     & 62.5 & \textbf{76.4} & 77.3 & 85.3 \\
Discussion   & 25.0 & 31.5 & 34.3 & 74.7 \\
\midrule
MAGRPO       & 74.3   & 76.0 & 77.5 & 85.3 \\
CoLLM-DC     & 59.1 & 60.5 & 60.5   & 62.8   \\
CoLLM-CC    & \textbf{75.2}   & 75.9 & \textbf{77.8}   & \textbf{86.5}   \\
\bottomrule
\end{tabular}
\vspace{2mm}
\caption{Pass@k results (1, 3, 5, 10) on \textit{CoopHE}. \textbf{Bolds} indicate the best performance. Results are averaged over 5 runs.}
\label{tab:coophe_passk}
\end{table}

Table~\ref{tab:coophe_passk} presents the pass@k performance of coding collaboration on \textit{CoopHE}. Fine-tuning with GRPO and AC yields marginal improvements over the raw model. However, this improvement over the given model is not due to acquired algorithmic knowledge or increased capacity. Instead, the training primarily refines the model’s policy, increasing the likelihood of producing correct solutions that already lie within its representational scope \cite{yue2025does}.

Most prompt-based multi-agent approaches underperform single-model baselines without proper optimization. Agents do not have timely communication to reason about each other’s correctness or functionalities. Although this challenge can be overcome through sequential execution, it reduces inference speed because agents take turns responding. 

MARL methods can achieve performance comparable to or better than that of a single larger model. CoLLM-CC consistently achieves the best results across most pass@k results, because the main agent is guided to infer the auxiliary agent's functionality and provide fallback solutions when the auxiliary utilities are vulnerable, e.g., the main can provide a boundary condition manipulation. Also, through MARL optimization, the main completes the remaining components of the implementation, thereby accelerating inference through parallel execution. CoLLM-DC performs worse than CoLLM-CC and MAGRPO due to non-convergence (Fig.~\ref{subfig:coophe}).

\subsection{Training Overhead}

We use the setting of \textit{CoopHE} and \textit{StrBuild} in Appendix~\ref{app:hyper} as a representative to compare MARL training overhead in Table~\ref{tab:cost_by_domain} and Table~\ref{tab:compute_costs}.

\begin{table}[H]
\setlength{\tabcolsep}{5pt}
\centering
\begin{tabular}{lccc}
\toprule
\textbf{Metric} & \textbf{MAGRPO} & \textbf{CoLLM-DC} & \textbf{CoLLM-CC} \\
\midrule
\#Epochs   & 8 & 80 & 80 \\
\#Rollouts   & 16 & 2 & 2 \\
\midrule
\#Samples  & 9640 & 8592 & 8438 \\
\#Updates   & 603 & 2148 & 2110 \\
Duration  & 4.5 & 13.4 & 11.1 \\
VRAM  & 93.8 & 126.3 & 107.4 \\
\bottomrule
\end{tabular}
\vspace{2mm}
\caption{Training overhead of MAGRPO, CoLLM-DC, and CoLLM-CC on \textit{CoopHE}, under the settings of Appendix~\ref{app:hyper}. Metrics include the number of epochs, rollouts per episode, total samples used, policy updates, training duration (\textit{H200} hours), and VRAM (GB) usage. Results are averaged over 5 runs.}
\label{tab:cost_by_domain}
\end{table}

Since MAGRPO is an instance of $K$-sampling, the rollout in an $H$-horizon episode forms a $K$-ary tree and satisfies Proposition~\ref{prop:marfsampleefficiency}. Under the hyperparameter setting in Appendix~\ref{app:hyper}, with $K=4$ and $H=2$, MAGRPO produces $K^H=16$ rollouts and uses $\sum_{k=1}^{H}K^H=20$ samples, assuming no early termination is triggered. In contrast, both CoLLM-DC and CoLLM-CC generate a single rollout consisting of $H=2$ samples. To ensure a comparable number of training samples for MARL, we therefore train CoLLM-DC and CoLLM-CC for $10$ times as many epochs as MAGRPO. Also, MAGRPO operates with a larger effective minibatch size. Accordingly, the agent's learning rate is designed to scale proportionally, resulting in fewer updates.

Training CoLLM-DC and CoLLM-CC requires substantially longer time and more GPU memory than MAGRPO, as critic LLMs must be maintained throughout training. However, this overhead does not scale linearly with the number of LLMs in the systems ($126.3 \ll 93.8 \times 2$). This is because the critic provides lower-variance gradient estimates, thereby reducing the number of samples that need to be retained in GPU memory for back-propagation. Furthermore, CoLLM-CC employs a single centralized critic, whose update cost is lower than maintaining n independent critics of the same size in CoLLM-DC, thereby reducing training time. As CoLLM-CC converges faster (Fig.~\ref{subfig:coophe}) and a lot of samples trigger early termination during the early stages of training, it ultimately uses the fewest samples.

\begin{table}[H]
\setlength{\tabcolsep}{3pt}
\centering
\resizebox{\linewidth}{!}{
\begin{tabular}{lcccc}
\toprule
\textbf{Metric} & \textbf{MAGRPO} & \textbf{CoLLM-DC} & \textbf{CoLLM-DC$^\ddagger$} & \textbf{CoLLM-CC} \\
\midrule
\multicolumn{5}{c}{$n=2$} \\
\midrule
\#Epochs   & 16 & 120 & 120 & 120 \\
\#Samples  & 6470 & 6264 & 6336 & 5620 \\
Duration   & 48h (1$\times$H200) & 52h (4$\times$H100) & 18h (1$\times$H200) & 20h (1$\times$H200) \\
VRAM       & 98G & 188G & 126G & 145G \\
\midrule
\multicolumn{5}{c}{$n=3$} \\
\midrule
\#Epochs   & 6 & -- & 80 & 80 \\
\#Samples  & 3488 & -- & 6929 & 5571 \\
Duration   & 68h (2$\times$H200) & -- & 43h (2$\times$H200) & 46h (2$\times$H200) \\
VRAM       & 167G & -- & 219G & 263G \\
\bottomrule
\end{tabular}
}
\vspace{2mm}
\caption{Training overhead of MAGRPO, CoLLM-DC, CoLLM-DC with shared actor-critic parameters (CoLLM-DC$^\ddagger$), and CoLLM-CC on \textit{StrBuild}, under the settings of Appendix~\ref{app:hyper}. Metrics include the number of training epochs, total samples used, training duration, and VRAM usage.}
\label{tab:compute_costs}
\end{table}

Table~\ref{tab:compute_costs} demonstrates the training overhead as the number of agents increases. MAGRPO incurs rapidly increasing training time as the number of agents grows, making it difficult to scale to larger MAS. CoLLM-DC faces a similar issue, since each agent requires its own actor and critic. Although CoLLM-DC$^\ddagger$ mitigates this overhead through parameter sharing, decentralized training remains susceptible to non-stationarity. CoLLM-CC avoids additional inference-time overhead under decentralized execution and achieves more stable training by using a centralized critic, but its training cost still increases with the MAS size. Developing more efficient and scalable training algorithms for multi-agent LLM systems remains an important future direction.

\section{Prompt Design Details} \label{app:prompt-design}

\subsection{Writing Collaboration}

In the \textit{TLDR} summarization, the instructions for each agent are as follows.

\begin{lstlisting}[numbers=none, escapeinside={(*}{*)}]

(*\textbf{Summary Agent}*) 
Create a concise summary response to this post.
Query: {prompt}
Instructions: Provide a brief and focused summary in a few sentences

(*\textbf{Elaboration Agent}*) 
Create a detailed summary response to this post.
Query: {prompt}
Instructions: You should use transition words to improve flow
\end{lstlisting}

In the \textit{arXiv} expansion, we use the \texttt{abstract} field of the dataset and process it as follows.

\begin{lstlisting}[numbers=none, escapeinside={(*}{*)}]

(*\textbf{Background Agent}*) 
Based on the following scientific abstract, expand the content for the introduction section.
Abstract: {abstract}
Instructions:
- There is another agent that will provide the method and implications
- You just need to focus on the background and motivation
- Avoid repeating methodology and implications content

(*\textbf{Method Agent}*) 
Based on the following scientific abstract, expand the content for the introduction section.
Abstract: {abstract}
Instructions:
- There is another agent that will provide the background and motivation
- You just need to focus on the method and implications
- Avoid repeating background and motivation content
\end{lstlisting}

\subsection{Coding Collaboration}

For CoopHE, we extract the \texttt{entry\_point}, \texttt{params} from the \texttt{prompt} field and instruct the agents as follows.

\begin{lstlisting}[numbers=none, escapeinside={(*}{*)}]

(*\textbf{Auxiliary Agent}*) 
Create a helper function for this coding problem.
Problem: {prompt}
Instructions:
- Output ONLY the function code, no explanations or examples
- Do NOT include markdown code blocks (```python)
- Do NOT include any text before or after the function
- Do NOT include test cases or example usage
- Create a helper function named 'aux' that can assist the main function
- The function should return useful data for solving the problem

Your output should follow this format:
def aux(...):
    # your code here
    return result
    
(*\textbf{Main Agent}*) 
Solve this coding problem by implementing the required function.
Problem: {prompt}
You have access to a helper function: aux(...)

Instructions:
- Output ONLY the function code, no explanations or examples
- Do NOT include markdown code blocks (```python)  
- Do NOT include any text before or after the function
- Do NOT include test cases or example usage
- Do NOT redefine the aux() function
- Implement ONLY the '{entry_point}' function as specified
- You can call aux() to assign a value to a variable within your function if helpful

Your output should follow this format:
def {entry_point}({params}):\n # your function code here\nreturn result\n
\end{lstlisting}

To improve the generated code, agents receive additional feedback in addition to the initial problem description. This feedback comes from the static analyzer and sandbox tests, and is appended to the prompts for subsequent turns.

\begin{lstlisting}[numbers=none, escapeinside={(*}{*)}]

Static and execution diagnostics:
- Main definition: FOUND (Main function prime_fib defined)
- Syntax: OK (Combined code syntax OK)
- Tests: 4/5 passed
assert candidate(1) == 2 
AssertionError: expected 2, got 1

Revise your prime_fib accordingly.

\end{lstlisting}

\subsection{Minecraft}

In \textit{StrBuild}, each agent is instructed by the following prompts at the first turn. The target specifications, available block textures, and building boundaries are provided as \texttt{target\_ascii}, \texttt{block\_agent\_lines}, and \texttt{{world\_bbox\_from}} and \texttt{world\_bbox\_to} to the agent, respectively.

\begin{lstlisting}[numbers=none, escapeinside={(*}{*)}]

You are Player i in an n-person Minecraft building team. You will place SOME of the blocks for the final build. Output must be Minecraft commands only (no markdown, no code fences, no extra text).

The target is a character grid made of '#' and '.'. Its size matches the bbox and target rows. Output /setblock commands for a subset of '#' positions.

Target grid (top-down rows): {target_ascii}

WORLD bbox (inclusive):
  - from: {world_bbox_from}
  - to:   {world_bbox_to}

Coordinate mapping (absolute coords):
  - Let (min_x, min_y, min_z) be bbox.from and (max_x, max_y, max_z) be bbox.to.
  - x increases from min_x to max_x (left to right).
  - y increases from min_y to max_y (bottom to top).
  - z is constant (min_z == max_z)

Available blocks (use ONLY these):
  {block_agent_lines}

Constraints:
  - Output ONLY Minecraft commands, one per line.
  - Allowed commands: /setblock only.
  - Use absolute integer coordinates only (no ~).
  - Place blocks ONLY at '#' positions; leave '.' as air.
  - Adjacent blocks (sharing a side) must NOT be the same texture.
  - Every coordinate must be within the bbox.

Format: /setblock <x> <y> <z> <block>

\end{lstlisting}

In \textit{HouseBuild}, since the player can construct blocks flexibly at any adjacent location, the positions of players are assumed to be the center. We employ a simulator to generate \texttt{spider\_num} spiders from arbitrary locations on the map. It also computes the player's health points \texttt{player\_hp}, and the damage inflicted on spiders \texttt{spider\_dmg}. The predefined player attack value is \texttt{player\_atk}, while each spider is assumed to have an attack value \texttt{spider\_atk}.

\begin{lstlisting}[numbers=none, escapeinside={(*}{*)}]

You are Player i in an n-person Minecraft building team. You will place SOME of the blocks for the final build. Output must be Minecraft commands only (no markdown, no code fences, no extra text).

Task: Build the 3D structure from the provided y-axis slices.

Available blocks (use ONLY these): {block_agent_lines}

Layers (ascending WORLD y). Each layer is a set of rectangles in WORLD (x,z) coords:
  - Format: {(x1, y1, z1, x2, y2, z2 block_id), (x1, y1, z1,
  x2, y2, z2 block_id)}

WORLD bbox (inclusive):
  - from: {world_bbox_from}
  - to:   {world_bbox_to}

Threat check: There are {spider_num} spiders nearby; you have {player_hp} HP. Spider attack list: {spider_atk}, total damage dmg={spider_dmg}. If you want to kill them, output the attack command: /damage @e[type=spider,limit=1] {player_atk}
minecraft:player attack. All other commands should be normal building commands.

Constraints:
  - Output ONLY Minecraft commands, one per line.
  - Allowed commands: /fill and /kill
  - Fill format: /fill x1 y1 z1 x2 y2 z2 block
  - Use absolute integer coordinates only (no ~).
  - Use ONLY blocks from the legend.
  - Every coordinate must be within the bbox.
\end{lstlisting}

And we provide DC and CC with the following global information, $m_t$, during training.

\begin{lstlisting}[numbers=none, escapeinside={(*}{*)}]

You are at turn = {turn} over a {num_turns}-turn episode. 
Your last-turn reward is {last_reward} / {reward_max}.

\end{lstlisting}

\section{Reward Design Details} \label{app:reward-design}

Reinforcement Learning with Verifiable Rewards (RLVR) provides objective and reliable training signals. However, most RLVR methods in a multi-turn setting rely on terminal rewards, resulting in extremely sparse feedback that complicates credit assignment across multiple agents. Recent work addresses this by introducing process-based or intermediate rewards, often instantiated as manually designed rubrics \cite{uesato2022solving, duverifier, wu2025collabllm}. However, these approaches typically require substantial manual intervention for rubric engineering. It remains unclear whether such designs are still generalizable, consistent, and aligned with the intended objectives.

We provide verifiable reward signals at each turn, enabling fine-grained and objective supervision throughout the interaction. However, the cumulative return becomes dependent on the horizon, resulting in a dynamic return scale that must be carefully handled in policy optimization. We solve this problem by providing the critic with global information $m_t$, as shown in critic prompts in Appendix~\ref{app:prompt-design}.

\subsection{Writing Collaboration}

We evaluate \textit{TLDR} summarization along three dimensions: structural quality, style consistency, and logical coherence. Structural wellness measures the relative completion length and lexical diversity (unique-word ratio excluding stopwords). Full rewards are assigned when the length ratio lies in 1.6-3.2$\times$, and the unique-word ratio is $\geq$2.0$\times$; values within broader tolerances (1.1-5.0$\times$ for length and 1.3--2.0$\times$ for uniqueness) receive proportionally scaled rewards, while out-of-range cases incur zero reward and early termination. Style consistency is quantified by the Jaccard similarity of vocabularies between completions (excluding stopwords), capped at $0.03$ to balance lexical consistency with vocabulary expansion. Logical coherence is evaluated through transition-word usage in the combined text across 12 functional categories, with rewards scaled by category diversity as $\min(0.6\log(\mathrm{\#categories}+1), 1)$.

For arXiv expansion, we assess structural wellness, style consistency, and logical coherence by comparing the second paragraph against the first. Structural wellness rewards optimal length ratios of 1.0-1.3$\times$ and unique-word ratios of 0.7-1.3$\times$, assigns proportional rewards within acceptable ranges (0.8-1.5$\times$ for length and 0.5-1.7$\times$ for uniqueness), and terminates evaluation otherwise. Style consistency is measured using Jaccard similarity between the two completions, capped at 0.23 and normalized as the reward. Logical coherence is evaluated through transition-word usage across the same 12 categories, with rewards scaled by category diversity as $\min(0.4\log(\mathrm{\#categories}+1), 1)$.

\subsection{Coding Collaboration}

We evaluate coding collaboration along four dimensions: structural integrity, syntactic correctness, test pass rate, and cooperation quality. Structural integrity verifies that both auxiliary and main functions are properly defined, with valid signatures and return statements; failure to define the main function results in immediate termination. Syntactic correctness assesses whether the concatenated code parses without errors, verified via the Abstract Syntax Tree (\textit{AST}) library. Any syntax error leads to evaluation termination to prevent runtime failures. The test pass rate measures the proportion of unit tests that successfully pass within an 8-second timeout, with rewards scaled by the number of successful assertions (tests) $\frac{\mathrm{\#passed\_tests}}{\mathrm{\#total\_tests}}$, and termination triggered if no tests pass. Cooperation quality assigns a base bonus when the main function invokes the auxiliary, with additional rewards granted when the main function exhibits substantive logic beyond a simple wrapper. A deduction is applied if the main function calls the auxiliary but discards its return value, penalizing superficial cooperation.

\subsection{Minecraft}

In \textit{StrBuild}, we first design the reward to encourage agents to construct string-like structures in accordance with the given paradigms using coverage rate $\frac{2 \mathrm{\#covered\_blocks}}{\mathrm{\#total}}$. The agents are also optimized to minimize unnecessary resource consumption $-\frac{1.5\mathrm{\#extra\_blocks}}{\mathrm{\#total}}$. In addition, we introduce an adjacency penalty based on the same-texture adjacency rate, $\frac{\mathrm{\#same\_texture\_pairs}}{2\mathrm{\#total\_pairs}}$, which discourages excessive adjacency of identical block textures. In \texttt{HouseBuild}, we similarly employ the coverage rate and redundancy rate to encourage accurate and efficient construction. Also, agents are penalized for damage inflicted by spiders as $\min\!\left(1, \frac{\mathrm{spider\_total\_dmg}}{\mathrm{player\_hp}}\right) \times 0.2$.

\section{Compute Resources} \label{app:compute}

The compute resources used in our experiments (Sec.~\ref{sec:exp}) for training and evaluation are listed below.

\subsection{Training Devices}

\begin{itemize}
  \item \textbf{\textit{Writing Collaboration}}
    \begin{itemize}
      \item Type: GPU Cluster \\
            CPU: NVIDIA Grace ARMv9 \\
            GPU: 1$\times$ NVIDIA Hopper H100
    \end{itemize}

  \item \textbf{\textit{Coding Collaboration}}
    \begin{itemize}
      \item Type: GPU Cluster \\
            CPU: Intel Xeon Platinum 8558 \\
            GPU: 1$\times$ NVIDIA Hopper H200

      \item Type: GPU Cluster \\
            CPU: Intel Xeon Gold 5318Y \\
            GPU: 1$\times$ NVIDIA Hopper H200

      \item Type: GPU Cloud Instance \\
            CPU: AMD EPYC 9755 \\
            GPU: 1$\times$ NVIDIA Hopper H200

      \item Type: GPU Cloud Instance \\
            CPU: Intel Xeon Platinum 8568Y+ \\
            GPU: 1$\times$ NVIDIA Hopper H200
    \end{itemize}

  \item \textbf{\textit{Minecraft Building Collaboration}}
    \begin{itemize}
      \item Type: GPU Cluster \\
            CPU: Intel Xeon Platinum 8592+ \\
            GPU: 1$\times$ NVIDIA Blackwell B200

      \item Type: GPU Cluster \\
            CPU: Intel Xeon Platinum 8558 \\
            GPU: 1$\times$ NVIDIA Hopper H200

      \item Type: GPU Cloud Instance \\
            CPU: Intel Xeon Platinum 8568Y+ \\
            GPU: 1$\times$ NVIDIA Blackwell B200

      \item Type: GPU Cloud Instance \\
            CPU: AMD EPYC 9655 \\
            GPU: 1$\times$ NVIDIA Hopper H200

      \item Type: GPU Cloud Instance \\
            CPU: AMD EPYC 9755 \\
            GPU: 1$\times$ NVIDIA Hopper H200

      \item Type: GPU Cloud Instance \\
            CPU: Intel Xeon Platinum 8592+ \\
            GPU: 1$\times$ NVIDIA Hopper H200
    \end{itemize}
\end{itemize}

\subsection{Inference Device}

\begin{itemize}
  \item \textbf{\textit{All Tasks}}
    \begin{itemize}
      \item Type: Standalone Workstation\\
            CPU: AMD Ryzen 9 9950X\\
            GPU: 1$\times$ NVIDIA GeForce RTX 5090
    \end{itemize}
\end{itemize}

\section{Code and Dataset} \label{app:codedata}

Our code and dataset are available as follows.

\begin{itemize}
    \item \texttt{CoMLRL}: \textbf{C}ooperative \textbf{M}ulti-\textbf{L}LM \textbf{R}einforcement \textbf{L}earning (CoMLRL) is an open-source library for training multiple LLMs to collaborate using Multi-Agent Reinforcement Learning (MARL). It provides implementations of various MARL trainers for decentralized LLM collaboration, {\url{https://github.com/OpenMLRL/CoMLRL/releases/tag/v1.3.6}}. Please refer to the documentation at \url{https://github.com/OpenMLRL/CoMLRL} for more detailed usage examples. 

    \item \texttt{LLM\_Collab\_Writing}: This repository contains our writing collaboration experiments, {\url{https://github.com/OpenMLRL/LLM_Collab_Writing/releases/tag/v1.3.6}}.

    \item \texttt{LLM\_Collab\_Code\_Generation}: This repository contains coding collaboration experiments, {\url{https://github.com/OpenMLRL/LLM_Collab_Code_Generation/releases/tag/v1.3.6}}.

    \item \texttt{LLM\_Collab\_Minecraft}: This repository contains Minecraft game-playing experiments, {\url{https://github.com/OpenMLRL/LLM_Collab_Minecraft/releases/tag/v1.3.6}}.

    \item \texttt{CoopHumanEval}: Our CoopHumanEval dataset we used is open-sourced at \url{https://huggingface.co/datasets/OpenMLRL/CoopHumanEval}.
    
\end{itemize}

\end{document}